\documentclass[lettersize,journal]{IEEEtran}
\usepackage{amsmath,amsfonts, bm}
\usepackage{algorithmic}
\usepackage{algorithm}
\usepackage{amssymb}
\usepackage{makecell}
\usepackage{amsthm} 
\usepackage{array}
\usepackage{balance}
\usepackage[caption=false,font=normalsize,labelfont=sf,textfont=sf]{subfig}
\usepackage{textcomp}
\usepackage{colortbl}
\usepackage{url}
\usepackage{pifont}
\usepackage{verbatim}
\usepackage{graphicx}
\usepackage{tabularray}
\usepackage{hyperref} % use this can add rectangle
\usepackage{units}
\usepackage{multirow}
\usepackage{cite}
\usepackage{soul, color}
\usepackage[dvipsnames, svgnames, x11names]{xcolor}
\soulregister{\cite}7
\soulregister{\citep}7
\soulregister{\citet}7
\soulregister{\ref}7
\soulregister{\pageref}7
\begin{document}

\title{Hierarchical Sparse Graph for Multi-dimensional Assignment in Event-based Tracking}

\author{Muxi Zha, Yang Shang, Wenlong Zhang, Qifeng Yu, and Banglei Guan
    \thanks{This work was supported by the Science and Technology Innovation Program of Hunan Province 2025RC1045, and National Natural Science Foundation of China 12372189 and 12202485.}
    \thanks{The authors are with the College of Aerospace Science and Engineering, National University of Defense Technology, Changsha, 410073, Hunan, China, and with the Hunan Provincial Key Laboratory of Image Measurement and Vision Navigation, Changsha, 410073, Hunan, China. (Corresponding authors: Yang Shang; Banglei Guan).}
    \thanks{Digital Object Identifier 10.1109/TCSVT.2026.3704671}
}

\IEEEpubid{\begin{minipage}{\textwidth} \centering Copyright~\copyright~2026 IEEE. Personal use of this material is permitted. \\ However, permission to use this material for any other purposes must be obtained from the IEEE by sending an email to pubs-permissions@ieee.org. \end{minipage}}
\maketitle

\begin{abstract}
    Existing event-based tracking methods typically operate in an online manner, associating tracklets across adjacent batches. However, this limited temporal window often leads to failures in challenging scenarios, such as long-term occlusions or missed detections, where broader temporal context should be incorporated for high-precision tracking. To achieve robust and accurate tracking, this paper concentrates on the event-based multi-dimensional assignment that associates tracklets across multiple batches simultaneously. Unlike image-based detectors, event-based detectors can generate multiple correct yet spatially distinct tracklets for a single object per batch. This violates the one-to-one rule in classic data association, making event-based association an underdetermined many-to-many matching problem. To the best of our knowledge, this is the first work to tackle this NP-hard problem. To address its underdetermined nature, we formulate the assignment problem via fuzzy logic, where pairwise tracklet correlations are quantified through multiple long-range kinematic metrics. This fuzzy design eliminates the need for hard tracking thresholds and enables multiple heterogeneous metrics to be integrated into a unified correlation. By treating tracklets as nodes and their fuzzy correlations as edges, we construct a Hierarchical Sparse Graph (HSG). Building upon the HSG, we pose assignment as a graph partition task, and employ a heuristic solution that measures vertex affinity for partitioning adaptively. The entire pipeline is training-free. Comprehensive experiments on two challenging event datasets, eTram and Ev-UAV, show that our method outperforms existing event-based and most image-based association models. Particularly, it achieves over 2$\times$ the performance of OC-SORT across all metrics in the many-to-many association scenario.

\end{abstract}


\section{Introduction}
\label{section:Intro}

\IEEEPARstart{D}{ata} assignment is a critical step in Multiple Object Tracking (MOT)~\cite{baeRobustOnlineMultiobject2014, gaoRobustTrackingLearning2023, rakaiDataAssociationMultiple2022, wangMambaEVTEventStreamBased2026}, especially in the tracking-by-detection paradigm. It is a classic combinatorial optimization problem aimed at identifying matching pairs among detections (or tracklets). Depending on the number of frames considered, image-based data association can be roughly categorized into two types: linear assignment for online tracking and multi-dimensional assignment for offline tracking. Although linear assignment methods~\cite{tokmakovLearningTrackObject2021, liuMultiObjectTrackingMeets2022, caoObservationCentricSORTRethinking2023} have shown strong performance, certain challenges, such as long-term occlusions or missed detections~\cite{zhangGlobalDataAssociation2008, linLTTrackRethinkingTracking2024}, can lead to a performance drop. To address this problem, a few works~\cite{buttMultitargetTrackingLagrangian2013, ochsHigherOrderMotion2012} incorporate multiple frames into the association process. These approaches are termed multi-dimensional assignment.

\begin{figure}[!t]
    \begin{center}
        \centerline{\includegraphics[width=1\linewidth]{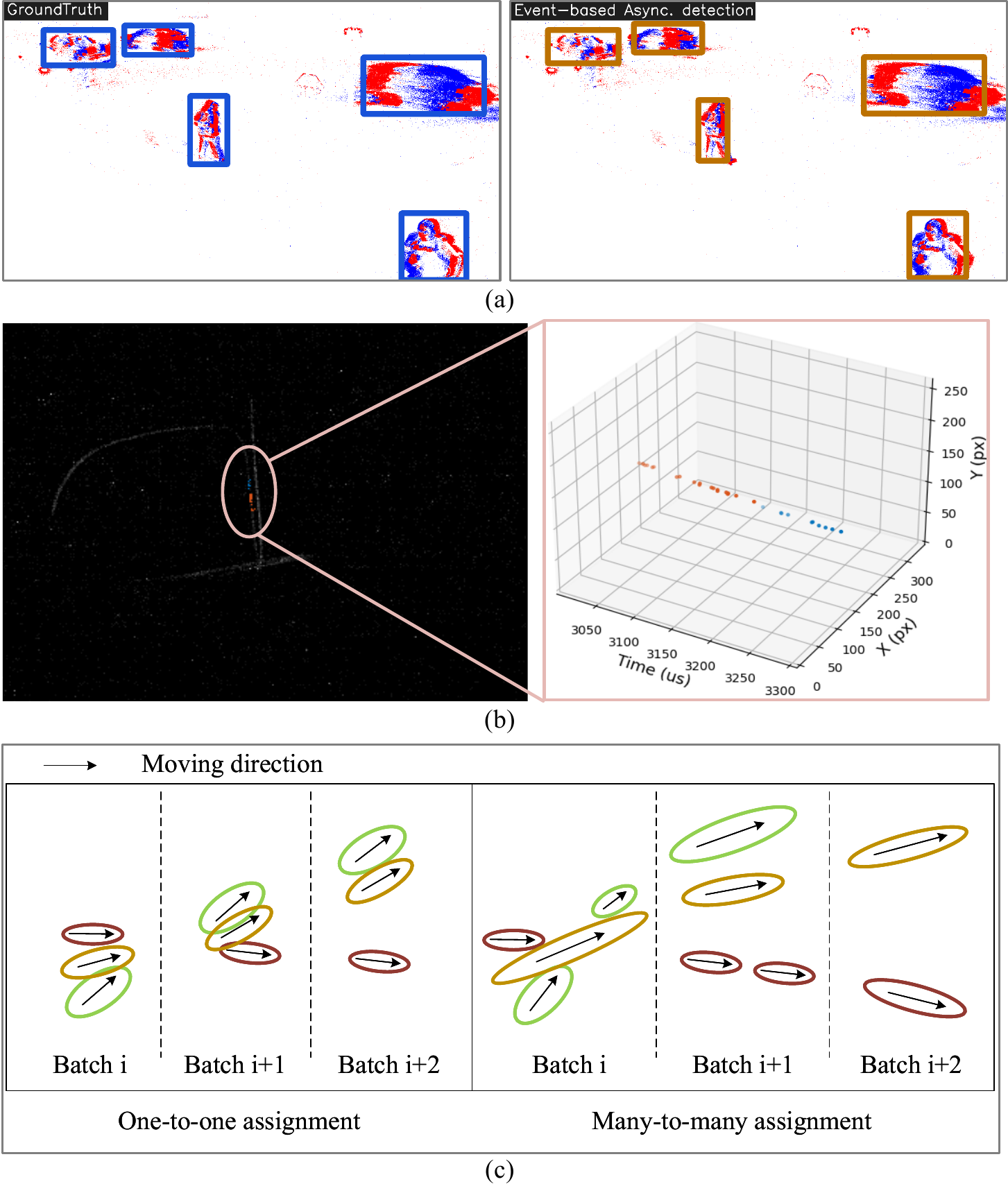}}
        \caption{Illustration of event-based detections and tracklet association. (a) Detection comparison between ground-truth (GT) and the state-of-the-art (SOTA) event-based detector RVT~\cite{gehrigRecurrentVisionTransformers2023} on eTram~\cite{vermaETraMEventbasedTraffic2024}. RVT produces detection boxes at a fixed rate, emulating a frame-like output. (b) 2D and 3D visualizations of tracklets within a 300 ms interval. Distinct colors represent different tracklets. These tracklets are obtained by the SOTA event-based segmentation network EV-SpSegNet~\cite{chenEventbasedTinyObject2025}, followed by region growing to cluster spatially connected events for association. (c) Comparisons of two types of event-based data association: image-like one-to-one association, as shown in (a), and event-based many-to-many association, as shown in (b).}
        \label{fig1_1}
        \vspace{-0.5cm}
    \end{center}
\end{figure}

\IEEEpubidadjcol

Event cameras are a notable class of bio-inspired sensors that work differently from conventional image-based cameras. Rather than outputting images at a fixed rate, they asynchronously capture per-pixel brightness changes as a stream of binary events. This unique working principle offers distinctive properties compared to image-based cameras, including high temporal resolution (1 $\mu\text{s}$), high dynamic range ($>$120 dB), and high sensitivity to dynamic stimuli~\cite{gallegoEventbasedVisionSurvey2022, huangEventGuidedStructuredOutput2018a}. These characteristics make event cameras particularly suitable for object tracking~\cite{wangPriorKnowledgeDrivenHybrid2025, liuEMTrackEfficientMultimodal2025, mengBRMoEBlindMultiModal2025} in challenging conditions, like fast motion, extreme illumination, or low-texture targets, where image-based cameras often fail.
In such scenarios, achieving high tracking accuracy and robust performance becomes the priority. This motivates a rethinking of the prevailing event-based efficiency-oriented tracking design and calls for more accurate event-based tracking approaches.

Event-based association is a critical tracking component but has received limited attention. In this paper, we study event-based multi-dimensional assignment to improve the overall tracking performance.
As illustrated in Fig.~\ref{fig1_1}(a) and (b), upstream detections lead to two distinct association scenarios. The former resembles image-based association, where each object corresponds to exactly one detection per batch. We refer to this as one-to-one association. But for the other scenario, this one-to-one characteristic does not hold since asynchronous events can yield multiple spatially distinct yet valid detections for a single object. While image-based detectors can also produce multiple detections per object, this matching ambiguity can be eliminated via Non-Maximum Suppression (NMS) or other strategies. In event-based vision, multiple detections for an object can all be correct and cannot be suppressed, as they may be different temporal segments of the event trajectory. In this case, event-based association becomes an underdetermined many-to-many matching problem. As the association period increases, the processing difficulty grows substantially.

\begin{figure*}[!t]
    \begin{center}
        \centerline{\includegraphics[width=1\linewidth]{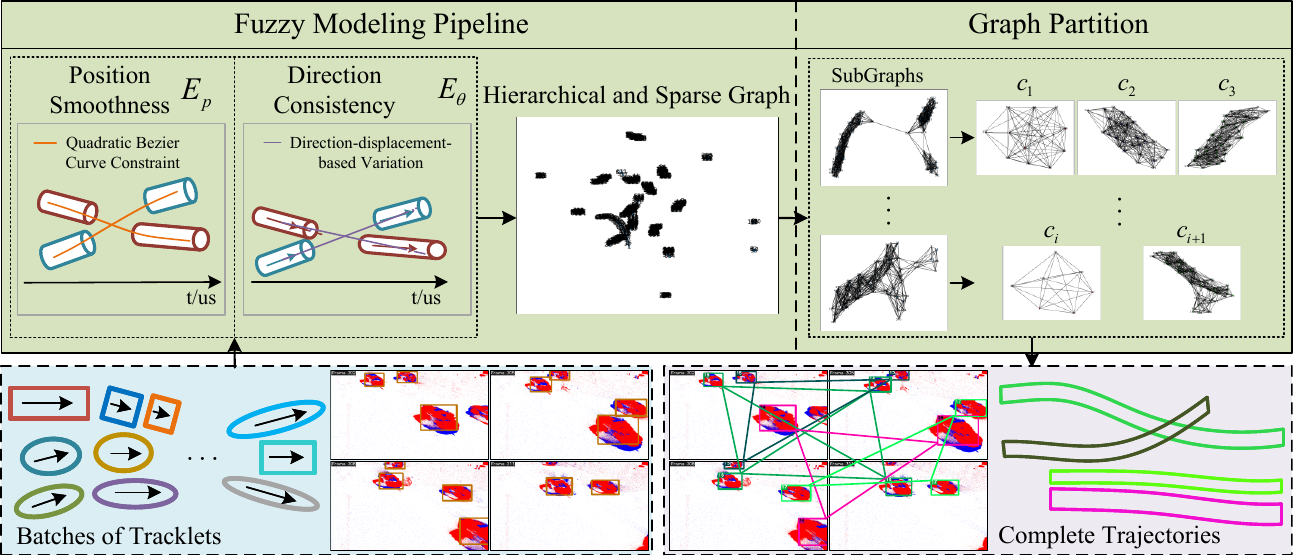}}
        \caption{Flowchart of the event-based multi-dimensional assignment optimization. Batches of tracklets from event-based detectors are fed into the fuzzy modeling pipeline. Positional and directional affinities are first computed to evaluate pairwise tracklet relationships. These metrics are integrated via hierarchical and sparse quantization to produce unified correlations, based on which the HSG is constructed. Finally, graph partition is performed on the HSG to identify cohesive partitions as optimal assignments.}
        \label{fig3_1}
    \end{center}
    \vspace{-.5cm}
\end{figure*}

Unlike the one-to-one bipartite matching with binary outcomes, event-based association is an inherently fuzzy problem due to the non-unique correspondence between objects and tracklets. Therefore, we construct a fuzzy modeling pipeline that takes two motion-based metrics as input and yields a fuzzy correlation value in $[0, 1]$ as output. Since no explicit mathematical mapping exists between these inputs and the output, we develop a hierarchical quantization method to bridge them, and subsequently propose a heuristic solution to seek the optimal assignment. The overall framework is shown in Fig.~\ref{fig3_1}.

The assignment problem is underdetermined, so we design a heuristic search-merge approach to find the optimal solution, which iteratively groups tracklets into cohesive partitions of HSG. Each partition corresponds to a set of tracklets of the same object. Central to this approach is a connection similarity measure, which leverages topological and weight information to assess the vertex affinity for merging. As this measure depends on multiple factors, manually setting a merging threshold is impractical. We derive an adaptive threshold under a null model to make the whole optimization automatic.
Our method is evaluated by comparing with multiple event-based and image-based association methods on eTram~\cite{vermaETraMEventbasedTraffic2024} and Ev-UAV~\cite{chenEventbasedTinyObject2025} datasets for the above two association cases.
Contributions are summarized as follows:
\begin{itemize}
    \item This is the first work on event-based multi-dimensional assignment. We model the problem via fuzzy logic to integrate multiple kinematic metrics into unified tracklet correlations and construct a HSG for assignment optimization.

    \item To model unpredictable and non-linear long-range event trajectories, we devise two kinematic metrics, B\'{e}zier curve-based position smoothness and direction-displacement-based variation consistency. These metrics capture the spatio-temporal coherence of tracklets in positional and directional domains, respectively.

    \item A hyperparameter-free graph partition solution is proposed. It quantifies vertex affinities and partitions the graph adaptively for optimal assignment.

    \item Despite being training-free, our method consistently outperforms existing event-based and motion-based methods, while remaining competitive with leading image-based association networks. Notably, it also demonstrates robust performance in tracking tiny objects (6.8 $\times$ 5.4 pixels) under large spatio-temporal gaps, surpassing OC-SORT~\cite{caoObservationCentricSORTRethinking2023} by over 2$\times$ across all metrics.
\end{itemize}

The remainder is as follows.
Section~\ref{section:Rew} reviews related work. Section~\ref{section:HSG} presents our fuzzy modeling pipeline and HSG construction, followed by the graph partition solution in Section~\ref{section:Solu}. Experimental results and analysis are reported in Section~\ref{section:exps}, and Section~\ref{section:Con} concludes the paper.

\section{Related works}
\label{section:Rew}
To date, event-based multi-dimensional assignment remains largely unexplored.
Existing event-based tracking methods often associate detections from adjacent batches or continuous timestamps for efficiency. Accordingly, we review both event-based association methods and relevant image-based multi-dimensional assignment techniques.

\subsection{Event-based association}
Existing event-based association methods can be broadly grouped into two paradigms, direct association and tracking-by-detection. Direct association methods~\cite{dietschePowerlineTrackingEvent2021, everdingLowLatencyLineTracking2018, legentilIDOLFrameworkIMUDVS2020} usually assign individual events to existing clusters, which is a unique paradigm benefiting from asynchronously triggered events. Tracking-by-detection is the classic paradigm in the image-based tracking domain. Therefore, most existing event-based association methods are derived from the image-based counterparts.
These methods typically employ Kalman Filter with the Hungarian algorithm~\cite{zhuEventBasedVisualInertial2017, jiangObjectTrackingEvent2020, wangEventBlobTracking2023, barranco2018real}, neighbor searching~\cite{alzugarayAsynchronousCornerDetection2018, lagorceAsynchronousEventBasedMultikernel2015, alzugarayHASTEMultiHypothesisAsynchronous2020, hu2022ecdt, rameshETLDEventbasedFramework2020}, or Intersection over Union (IoU)-based matching~\cite{chenAsynchronousTrackingDetectionAdaptive2019, chenEndtoEndLearningObject2020}. For instance, Alzugaray et al.~\cite{alzugarayHASTEMultiHypothesisAsynchronous2020} use a multi-hypothesis framework where the alignment score of each hypothesis is updated incrementally to track features in adjacent batches.
Hu et al.~\cite{hu2022ecdt} generate head and tail descriptors from each tracklet, then match head descriptors of the previous batch with tail descriptors of the current batch.
Jiang et al.~\cite{jiangObjectTrackingEvent2020} use an offline-trained detector and an online-trained
tracker, where a Kalman fusion scheme has been proposed to incorporate measurements from the detector and the tracker for real-time intervention-free operations.
Barranco et al.~\cite{barranco2018real} utilize Mean Shift to cluster event tracklets in X-Y-T space and employ Kalman Filter with Hungarian algorithm to associate them in adjacent batches.
Chen et al.~\cite{chenAsynchronousTrackingDetectionAdaptive2019} detect objects by bounding boxes in event frames and use IoU to evaluate association candidates. Despite their diversity in implementation, all these methods operate on adjacent batches, making them unable to associate fragmented tracklets with large spatio-temporal gaps.

\subsection{Image-based multi-dimensional assignment}
A fundamental constraint in image-based association is the one-to-one constraint. It states that each object must correspond to exactly one detection and one detection must belong to only one object. This results in a bijective mapping between detections and objects. In image-based multi-dimensional assignment, although the increasing number of candidates elevates the complexity of the problem, this constraint renders this NP-hard problem well-posed.

Some works have formulated multi-dimensional assignments as a network flow problem~\cite{berclazMultipleObjectTracking2011, zhangGlobalDataAssociation2008, pirsiavashGloballyoptimalGreedyAlgorithms2011, collinsMultitargetDataAssociation2012, poschmannFactorGraphBased2020}, with nodes as detections and edge weights based on appearance or motion.
Butt et al.~\cite{buttMultitargetTrackingLagrangian2013} treat tracking as a binary linear program and apply Lagrangian Relaxation to decompose the assignment into a series of simpler min-cost network flow subproblems.
Collins et al.~\cite{collinsMultitargetDataAssociation2012} break the problem into several two-frame linear assignments solved with the Hungarian algorithm. While these approaches yield good results, some works explore diverse modeling methods for the multi-dimensional assignment problem.
Poschmann et al.~\cite{poschmannFactorGraphBased2020} avoid explicit data association. They model association as a Gaussian Mixture Model (GMM) within a factor graph.
Ochs et al.~\cite{ochsHigherOrderMotion2012} build a hypergraph and use spectral clustering to obtain trajectories.
Roshan et al.~\cite{roshanzamirGMCPTrackerGlobalMultiobject2012a} formulate data association as a Generalized Minimum Clique Problem (GMCP) on a graph. This method addresses one object's data association at a time and employs Tabu Search to find the clique that minimizes a global cost combining appearance similarity and motion consistency iteratively.
Based on this work, Dehghan et al.~\cite{dehghanGMMCPTrackerGlobally2015} model a Generalized Maximum Multi Clique Problem (GMMCP) and use Binary Integer Programming (BIP) to find K-cliques that maximize the sum of edge weights with dummy nodes to handle occlusions.

While image-based methods work for the one-to-one association, they still fail in the many-to-many association. Besides, these methods typically utilize appearance cues for association, but such information is largely unavailable in events. We therefore rethink the modeling of event-based multi-dimensional assignment and leverage only kinematic features to capture relationships between tracklet pairs.

\section{Fuzzy-based optimization formulation}
\label{section:HSG}

This section first presents the optimization objective for event-based multi-dimensional assignment. We then detail our fuzzy modeling pipeline, including the definition and computation of two input metrics, the hierarchical and sparse quantization, and the resulting HSG.

\subsection{Event-based Multi-dimensional Assignment Optimization}
\label{section:optim}

We formulate event-based multi-dimensional assignment as a graph-based partitioning problem. Let $G$ be the constructed HSG, where nodes are tracklets and edge weights are their correlations. The goal is to obtain a partition $C = \{c_i\}_{i=1}^{K}$ of $G$ that maximizes the sum of intra-partition weights $\varphi_c$ while minimizing inter-partition weights $\psi(c_i, c_j)$. Main difficulties of the optimization lie in the unknown number of partitions $K$, which denotes the number of targets, and the non-deterministic relationship between $K$ and the number of tracklets.

\begin{equation}
    \label{eq3_1}
    \begin{aligned}
        \mathop{argmax}\limits_{C} \quad & \sum_{c \in C} \varphi_c  - \sum_{\substack{c_i, c_j \in C, i \neq j \\ c_i \cap c_j = \emptyset}} \psi(c_i, c_j), \\
        \text{s.t.} \quad                & \bigcup_{c \in C} c = G.                                             \\
    \end{aligned}
\end{equation}

\subsection{B\'{e}zier curve-based Position Smoothness}
\label{section:PS}
Straight line motion is a common and effective assumption in event-based vision~\cite{zhu2017event, zhuEventBasedVisualInertial2017}, but it only holds over short time intervals. Over long durations, object trajectories often exhibit complex patterns due to unpredictable acceleration. To effectively associate long-range tracklets, we additionally introduce a position smoothness metric based on B\'{e}zier curves.

\begin{figure}[!t]
    \begin{center}
        \centerline{\includegraphics[width=1\linewidth]{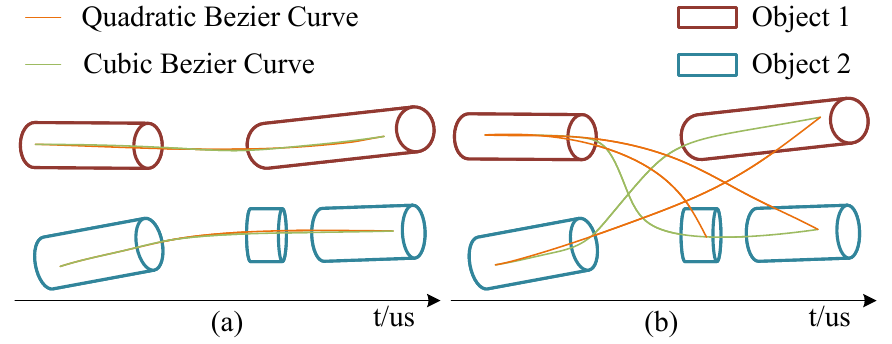}}
        \caption{Comparison of quadratic and cubic Bézier curve fitting. Trackets from (a) one object and (b) different objects.}
        \label{fig3_2}
    \end{center}
    \vspace{-.5cm}
\end{figure}

B\'{e}zier curve~\cite{seokRobustFeatureTracking2020a, chenAccurateEfficientApproximation2017} is a flexible polynomial representation capable of expressing complex curves with relatively few control points. The expressive power of a B\'{e}zier curve depends on its order, i.e., the number of control points. Higher-order curves can fit complex trajectories more accurately. However, they do not always improve matching reliability, since they tend to overfit the data with overly intricate curvature.
A comparison between the quadratic and cubic B\'{e}zier curve fitting, as shown in Fig.~\ref{fig3_2}, illustrates this problem. For quantitative results, please refer to Sec.~\ref{section:ablation}. Though the cubic B\'{e}zier curve fits trajectory as accurately as the quadratic one, it also fits incorrect matching pairs well. This overfitting risk is further amplified by the high temporal resolution of event cameras, which provide finer-grained temporal units for curve evolution.
Consequently, we adopt the quadratic B\'{e}zier curve as the position fitting polynomial. As in Eq.~\ref{eq3_2}, the curve has three control points, $P_0$, $P_1$, and $P_2$. The independent variable $t$ describes the progression of curve generation.

\begin{equation}
    \label{eq3_2}
    B(t) = (1-t)^2P_0 + 2t(1-t)P_1+t^2P_2.
\end{equation}

Given two tracklets, $L_i$ and $L_j$, control points are computed from a combined set of their centerline samples, denoted as $\{p_k\}_{k=1}^{2M}$, where each tracklet provides $M$ samples. These points can be obtained either by extracting centroids at uniform time intervals of tracklets, or by using the point-normal representation of the centerline. Assuming that $p_1$ and $p_{2M}$ are the temporally earliest and latest samples, the start and end control points are fixed as $P_0=p_1$ and $P_2=p_{2M}$. $P_1$ is computed via least squares:

\begin{equation}
    \label{eq3_3}
    P_1 = \left(A^T A\right)^{-1} A^T b,
\end{equation}
where,
\begin{equation}
    A   = \begin{bmatrix}
        2t_1(1-t_1) \\
        \vdots      \\
        2t_{2M}(1-t_{2M})
    \end{bmatrix},
\end{equation}

\begin{equation}
    b   = \begin{bmatrix}
        p_1 - (1-t_1)^2 P_0 - t_1^2 P_2 \\
        \vdots                          \\
        p_{2M} - (1-t_{2M})^2 P_0 - t_{2M}^2 P_2
    \end{bmatrix}.
\end{equation}

\begin{figure}[!t]
    \begin{center}
        \centerline{\includegraphics[width=1\linewidth]{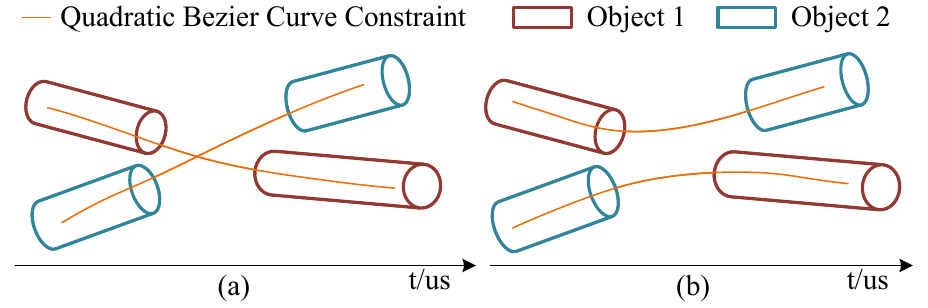}}
        \caption{Two association cases with both low $E_p$. (a) Correct and (b) incorrect matching pairs.}
        \label{fig3_3}
    \end{center}
    \vspace{-.5cm}
\end{figure}

$t$ consists of $2M$ values to fit all elements in $\{p_k\}_{k=1}^{2M}$. It naturally corresponds to the timestamps of these samples, reflecting how tracklet positions evolve over time.
B\'{e}zier curve-based position smoothness error is defined as the Mean Square Error (MSE) between the fitted points and samples, as given in Eq.~\ref{eq3_4_def2}.

\begin{equation}
    \label{eq3_4_def2}
    E_{p2} \triangleq \frac{1}{2M-2}\sum_{k=1}^{2M} \bigl(B(t_k)-p_k\bigr)^2.
\end{equation}

Unlike standard MSE, its denominator is $2M-2$ as fitting errors at endpoints are zero. Combining with straight line-based error $E_{p1}$, the final position smoothness error $E_p$ is

\begin{equation}
    \label{eq3_4}
    E_p \triangleq
    \begin{cases}
        E_{p1},~~~~ & E_{p1} \le \min(w_i, w_j) \\
        E_{p2},     & E_{p1} > \min(w_i, w_j)
    \end{cases},
\end{equation}

\begin{equation}
    \label{eq3_4_def1}
    E_{p1} \triangleq \min\bigl(\mathcal{D}(c_i, c_j, \mathbf{d_i}), \mathcal{D}(c_j, c_i, \mathbf{d_j})\bigr),
\end{equation}

\begin{equation}
    \label{eq3_4_H}
    \mathcal{D}(c_i, c_j, \mathbf{d_i}) \triangleq \bigl\lVert(\mathbf{c_i c_j} \times \operatorname{normalize}(\mathbf{d_i}))\bigr\rVert_2,
\end{equation}
where $w_i$ and $w_j$ are spatial widths of $L_i$ and $L_j$, calculated from detection boxes or estimated geometrically from event tracklets. $\mathbf{d_i}$ and $\mathbf{d_j}$ are unit direction vectors of $L_i$ and $L_j$, obtained similarly to the widths $w$. $\mathbf{c_ic_j}$ denotes the vector from centroid $c_i$ of $L_i$ to centroid $c_j$ of $L_j$. $\mathcal{D}(\cdot)$ is the distance function in X-Y-T space. $\Vert\cdot\Vert_2$ and $\text{normalize}(\cdot)$ are L2-norm and L2 normalization, respectively.

\subsection{Direction-displacement-based Variation Consistency}
\label{section:DS}

In the long-range tracklet association, B\'{e}zier curve-based position smoothness is an effective cue for continuous motion. However, it only serves to exclude tracklet pairs with implausible positional changes. When two tracklets are spatially close but exhibit significantly different velocities, this metric yields a small value, as illustrated in Fig.~\ref{fig3_3}. Similarly, our line-based error considers only positional changes. To further constrain tracklets of the same trajectory, we introduce a second cue for continuous motion: velocity consistency.

\begin{figure*}[!t]
    \begin{center}
        \centerline{\includegraphics[width=.96\linewidth]{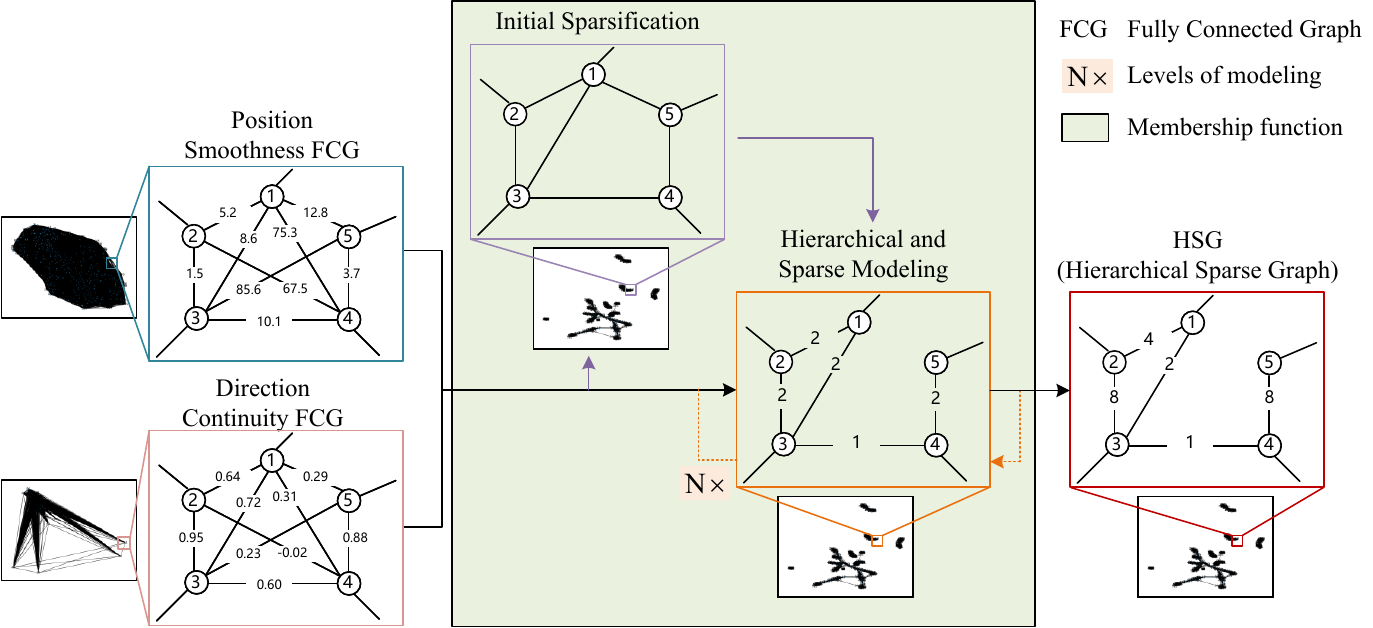}}
        \caption{Workflow of our fuzzy modeling pipeline. Compute two kinematic metrics for each pair of tracklets, and an intermediate FCG is generated for each metric. Since these metrics differ in orders and units, we perform hierarchical and sparse quantization on them to generate a unified correlation metric. Finally, the correlations and tracklets are represented as an HSG for subsequent processing.}
        \label{fig3_5}
    \end{center}
    \vspace{-.5cm}
\end{figure*}

\begin{figure}[!t]
    \begin{center}
        \centerline{\includegraphics[width=1\linewidth]{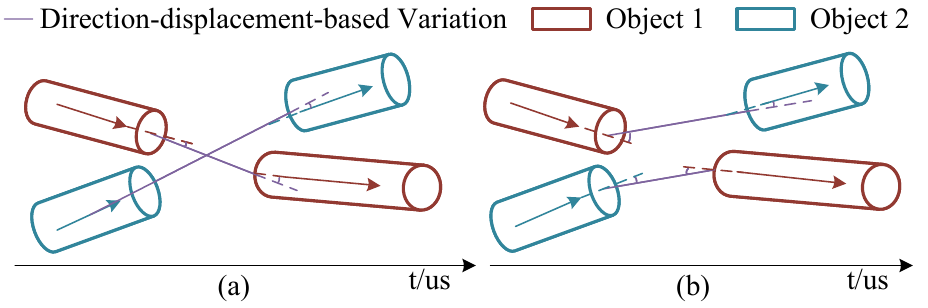}}
        \caption{Two association cases with direction continuity. (a) Correct matching pairs with big $E_\theta$, and (b) incorrect matching pairs with small $E_\theta$.}
        \label{fig3_4}
    \end{center}
    \vspace{-.5cm}
\end{figure}

Velocity can be decomposed into speed and direction. Since it is difficult to accurately estimate the speed of asynchronous events, we focus on the direction. When two tracklets originate from one object, the variation between their moving directions and positional displacement should be small. We quantify this variation as the average cosine similarity between the direction vectors of tracklets and the displacement vector:

\begin{equation}
    \label{eq3_5}
    E_\theta \triangleq \frac{\mathbf{d_i}\cdot \text{normalize}(\mathbf{c_ic_j}) + \text{normalize}(\mathbf{c_ic_j})\cdot\mathbf{d_j}}{2},
\end{equation}
where $L_i$ occurs earlier than $L_j$. Though $E_p$ is insufficient to distinguish correct associations shown in Fig.~\ref{fig3_3}, $E_\theta$ effectively disambiguates them by yielding higher values for correct matches, as illustrated in Fig.~\ref{fig3_4}.

\subsection{The Computation of Metrics}
\label{section:FCG}

The computation of two aforementioned metrics starts from temporally proximal candidates to distant ones. By representing tracklets as nodes and metric values as edge weights, we obtain two intermediate fully connected graphs (FCGs), one for each metric, as shown in Fig.~\ref{fig3_5}.
However, the high temporal resolution of events introduces two challenges during computation. First, $E_p$ becomes unreliable as temporal intervals increase. When the interval is large, two tracklets act like fixed anchors at the start and end of a long quadratic Bézier curve. This always yields a low $E_p$ even for incorrect matching pairs. Second, $E_\theta$ may lose its effectiveness in distinguishing correct associations. Event cameras offer microsecond precision, but a detection batch often spans hundreds of milliseconds. This duration is several orders of magnitude larger than the spatial scale (e.g., $1280\times 720$). As a result, direction vectors are dominated by the temporal component, making spatial variation negligible. Hence, $E_\theta$ remains large for all candidate pairs and may fail to differentiate correct matches from the rest.

Temporal scaling is a common strategy in event processing, typically implemented as exponential decay~\cite{lagorceHOTSHierarchyEventBased2016} to compress the temporal dimension. This strategy also balances the disparity in temporal and spatial scales, thereby helping to mitigate the second issue. However, it is not effective for the first one. Temporal scaling essentially samples the event stream along the t-axis. When a large scaling factor $s$ is applied, tracklets may degrade into points, further amplifying the first problem. To address it, we introduce a temporal weight factor $\lambda > 1$ to $E_p$ to adjust the association reliability with temporal intervals.

\subsection{Temporal Weight Analysis}
\label{sec:TWA}

$\lambda$ is introduced to reduce erroneous associations across large temporal gaps when computing positional error $E_p$. Theoretically, tracklets should be primarily associated with their temporal proximate neighbors. Only in the absence of such neighbors should the association be relaxed to include more temporally distant candidates. This association principle involves two factors, association interval $\Delta t$ and the historical match count $\alpha$. $\Delta t$ denotes the temporal gap between tracklets, which is an intuitive indicator of association reliability. If two tracklets are temporally distant, they are less likely to be associated. $\alpha$ reflects the number of times a tracklet has been successfully associated in the past. In our framework, $E_p$ computation proceeds from temporally proximal candidate pairs to more distant ones. For a tracklet with a high historical match count, further associations with remaining candidates should be treated with caution. Conversely, when a tracklet has low $\alpha$, its association reliability should remain high even as $\Delta t$ increases, allowing it to connect with more candidates and recover from fragmentation.

Besides, we incorporate object size $w$ to adjust the temporal weight for objects of different scales.
In dense or frequently overlapping scenarios, the positional errors of small objects tend to be small, while those of large objects tend to be large even after normalization. This is because small objects (e.g., pedestrians walking side by side) can be extremely close, resulting in negligible positional deviation. In contrast, large objects (e.g., vehicles) maintain larger separations, leading to larger normalized errors. To resolve the association ambiguity in such cases, we incorporate a scale-aware adjustment into the temporal weight $\lambda$, assigning higher weights to small objects and lower weights to large ones. The detailed determination of $\lambda$ and the effectiveness of each factor are reported in our supplementary material.

\subsection{Hierarchical and Sparse Quantization}
\label{section:HM}
Since no deterministic mapping exists between the input metrics and correlations, directly deriving the latter from the former is challenging. Furthermore, these metrics are heterogeneous in range and unit, where $E_p \in [0, \infty)$ is measured in pixels and $E_\theta \in [-1, 1]$ is unitless. This heterogeneity complicates their integration into a unified output.

\begin{figure}[!t]
    \begin{center}
        \vspace{-0.2cm}
        \centerline{\includegraphics[width=1.0\linewidth]{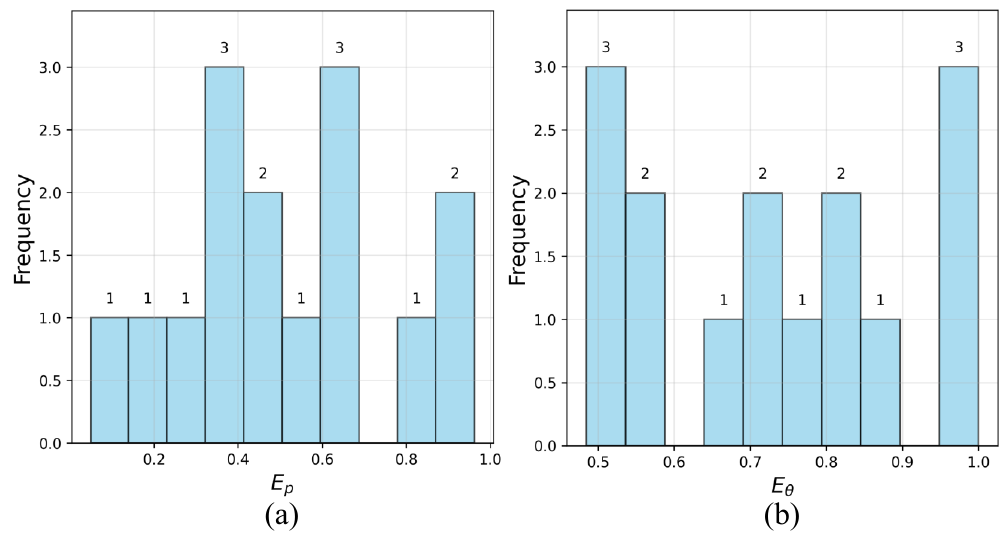}}
        \vspace{-0.4cm}
        \caption{Distribution of two metric values. (a) B\'{e}zier curve-based position smoothness error $E_p$, where lower values indicate stronger correlation. (b) Direction-displacement-based variation $E_\theta$, where higher values correspond to stronger correlation.}
        \label{fig3_6}
    \end{center}
    \vspace{-.5cm}
\end{figure}

Facing these issues, we construct the transformation between inputs and outputs based on an established principle that stronger correlation corresponds to lower $E_p$ and higher $E_\theta$.
As illustrated in Fig.~\ref{fig3_6}, distributions of $E_p$ and $E_\theta$ reveal clear gaps. These gaps typically arise from inherent correlation strengths or the temporal weight $\lambda$. We leverage these gaps to quantify continuous metrics into discrete correlations through hierarchical modeling. Sparse modeling is coupled to prune irrelevant edges, thereby reducing the computational burden.

Before hierarchical modeling, it is crucial to address the dependency of $E_p$ on object scale, since the same $E_p$ value implies different correlation strengths for objects of different scales. To mitigate this effect, we normalize $E_p$ by the object width $w$. Concurrently, we perform initial sparsification by filtering out pairs if their normalized positional error exceeds the object size ($E_p' > 1$) or direction-displacement variation angle exceeds $\nicefrac{\pi}{2}$ radians ($E_\theta < 0$). The remaining $E_p'$ and $E_\theta$ share the same range $[0, 1]$.

After initial sparsification, node $i$ has $j$ association candidates. We denote their positional errors as the set $\mathcal{E}_{p}^{'i}=\{E_{p}^{'i1}, \ldots, E_{p}^{'ij}\}$ and their direction-displacement variations as the set $\mathcal{E}_{\theta}^{i}=\{E_{\theta}^{i1}, \ldots, E_{\theta}^{ij}\}$. Hierarchical modeling assigns $N+1$ levels to these values, where levels $1$ to $N$ represent valid correlation levels, and level $0$ is reserved for outliers. We employ K-means++ with $N+1$ initial clusters. The cluster results are then sorted by centroids, descending for $E_p$ values and ascending for $E_\theta$ values, to assign levels from the lowest $0$ to $N$. Our method is not sensitive to clustering configurations, as evidenced by experimental results in the supplementary material.
In positional errors $E_p^{'}$, there are two fitting models. Since the line model imposes a stronger constraint than the B\'{e}zier curve model, we introduce a bias $\beta$ to differentiate their contributions in outputs. The final correlation $r^{ij}=r^{ji}$ between $i$ and $j$ is defined in Eq.~\ref{eq3_6}. It is the sum of directional correlation levels $l_1$ (from $i$ to $j$) and $l_2$ (from $j$ to $i$).

\begin{equation}
    \begin{small}
        \label{eq3_6}
        \begin{gathered}
            r^{ij} =
            \begin{cases}
                {l_1} + {l_2} + \beta , & \text{if } E_p^{'ij} \text{ is line-based } E_{p1}^{'ij}             \\
                {l_1} + {l_2}         , & \text{if } E_p^{'ij} \text{ is B\'{e}zier curve-based } E_{p2}^{'ij}
            \end{cases},
            \\
            l_1 =    \min(N_p^{ij}, N_\theta^{ij}) \in [0, N],                                                  \\
            l_2 =    \min(N_p^{ji}, N_\theta^{ji}) \in [0, N],                                                  \\
        \end{gathered}
    \end{small}
\end{equation}
where $N_p^{ij}$ and $N_p^{ji}$ denote the levels of $E_p^{'ij}$ and $E_p^{'ji}$, respectively. Similarly, $N_{\theta}^{ij}$ and $N_{\theta}^{ji}$ correspond to the levels of $E_{\theta}^{ij}$ and $E_{\theta}^{ji}$. The final output HSG is constructed on correlations. Detailed pseudocode of our fuzzy modeling pipeline is provided in Algorithm~\ref{alg:0}.

\begin{algorithm}[!t]
    \caption{Fuzzy Modeling Pipeline}
    \begin{small}
        \label{alg:0}
        \begin{algorithmic}[1]
            \REQUIRE Tracklet set $T$, hierarchical levels $N$.
            \STATE $G \gets \text{EmptyDirectedGraph}()$
            \STATE $HSG \gets \text{EmptyUndirectedGraph}()$\\
            \COMMENT{Calculate two metrics}
            \FOR{$i = 1$ \TO NumberOfNodes($G$)}
            \FOR{$j = i+1$ \TO NumberOfNodes($G$)}
            \STATE $E_{p1}^{ij}, E_{p2}^{ij} \gets \text{PosSmoothError}(T_i, T_j)$ using Eq.~\ref{eq3_4}
            \STATE $E_{p1}^{ij}, E_{p2}^{ij} \gets E_{p1}^{ij} \cdot \lambda_1, E_{p2}^{ij} \cdot \lambda_2$
            \STATE $E_\theta^{ij} \gets \text{DirectDisplaceVariation}(T_i, T_j)$ using Eq.~\ref{eq3_5}\\
            \COMMENT{Hierarchical \& sparse quantization: initial sparsification}
            \IF{$E_{p1}^{ij} < \min(w_i, w_j)$}
            \STATE $G \gets \text{AddEdge}(i, j, E_{p1}^{ij} / w_i, E_\theta^{ij})$
            \STATE $G \gets \text{AddEdge}(j, i, E_{p1}^{ij} / w_j, E_\theta^{ij})$
            \ELSIF {$E_{p2}^{ij} < \min(w_i, w_j)$}
            \STATE $G \gets \text{AddEdge}(i, j, E_{p2}^{ij} / w_i, E_\theta^{ij})$
            \STATE $G \gets \text{AddEdge}(j, i, E_{p2}^{ij} / w_j, E_\theta^{ij})$
            \ENDIF
            \ENDFOR
            \ENDFOR\\
            \COMMENT{Hierarchical \& sparse quantization: hierarchical \& sparse modeling}
            \FOR{$i$ in $\text{Nodes}(G)$}
            \STATE $\mathcal{E}_{p}^{'i}, \mathcal{E}_{\theta}^i \gets \text{GraphWeight}(G,~i)$
            \STATE $N_{p}^i, N_{\theta}^i \gets \text{HierarchicalSparseModel}(\mathcal{E}_{p}^{'i}, \mathcal{E}_{\theta}^i, N+1)$
            \STATE $HSG \gets \text{AddEdge}(i, N_{p}^i, N_{\theta}^i, \beta)$ using Eq.~\ref{eq3_6}
            \ENDFOR
            \RETURN $HSG$
        \end{algorithmic}
    \end{small}
\end{algorithm}

\section{Connection Similarity based Graph Partition}
\label{section:Solu}

Our optimization problem is an underdetermined combinatorial problem with no closed-form solution. Inspired by community detection in social network analysis~\cite{blondelFastUnfoldingCommunities2008, xuSCANStructuralClustering2007}, we adopt a heuristic search-merge solution.
Based on the topological structure and edge weights within HSG, we design a connection similarity with an adaptive threshold to automatically guide the vertex merging for graph partitioning.

\subsection{Connection Similarity}
\label{section:CS}

Connection similarity evaluates the vertex affinity for merging.
Consider two merged vertices $u_i=(V_i,R_i)$ and $u_j=(V_j,R_j)$, where a vertex consists of a few nodes $V$ and edges $R$. Their connection similarities are two asymmetric scores $S_i^j$ and $S_j^i$, as defined in Eq.~\ref{eq4_1}.
$S_i^j$ is the connection similarity from $u_j$ to $u_i$. It computes the proportion of edge weights shared between $u_i$ and $u_j$ via common neighbors relative to $u_i$'s total weights. $S_j^i$ is defined analogously.
This measure focuses on connectivity through shared neighbors rather than direct connections. Consequently, it allows merging vertices that share common neighbors even if they lack direct links.

\begin{equation}
    \label{eq4_1}
    \left\{
    \begin{aligned}
        \chi_{ij} & = \sum_{\substack{o\in V_i, p \in V_j          \\ k\in \textrm{Neighbor}(V_i)\bigcap \textrm{Neighbor}(V_j)}} \min(r^{ok}, r^{pk}) \\
        S_{i}^j   & = \frac{\chi_{ij}}{\sum_{\substack{o\in V_i,   \\ k\in\textrm{Neighbor}(V_i)}} r^{ok}} = \frac{\chi_{ij}}{S_i} \in [0, 1]\\
        S_{j}^i   & = \frac{\chi_{ij}}{\sum_{\substack{p\in V_{j}, \\ k\in \textrm{Neighbor}(V_j)}} r^{pk}} = \frac{\chi_{ij}}{S_j}\in [0, 1]
    \end{aligned},
    \right.
\end{equation}
where $\textrm{Neighbor}(\cdot)$ denotes the neighboring node set, $\chi_{ij}$ is the total edge weight shared between $u_i$ and $u_j$ via common neighbors, and $S_i$ and $S_j$ are the total edge weights of $u_i$ and $u_j$ respectively.

\subsection{Threshold for Connection Similarity under a Null Model}
\label{section:CSB}

Connection similarity depends on multiple factors, such as topological structures and edge weights, making it impractical to set a reliable threshold manually for merging decisions. Here, we provide a threshold for automatic decisions. Inspired by the null model used in modularity~\cite{newmanModularityCommunityStructure2006}, we derive an adaptive threshold $\tau$. This threshold approximates the expected connection similarity in a randomly connected graph, and serves as the lower bound for our vertex merging. Specifically, $u_i$ and $u_j$ are merged if $S_{i}^j \ge \tau_i$ or $S_j^i \ge \tau_j$.

To derive $\tau_i$, we first estimate the expected common edges shared between $u_i$ and $u_j$. For vertex $u_i$, let $n_v^i$ denote its number of nodes, $n_r^i$ the number of intra-vertex edges, $\zeta_i$ the average number of edges per node, and $r_i$ the average edge weight.
The internal graph density of $u_i$ is $D_i = \nicefrac{2n_r^i}{n_v^i(n_v^i-1)}$, and similarly for $u_j$, $D_j = \nicefrac{2n_r^j}{n_v^j(n_v^j-1)}$. Suppose $u_i$ has $m$ merging candidates that are similar to $u_j$, and the graph density of the null model consisting of $u_i$ together with $m$ candidates is estimated as

\begin{equation}
    \label{eq4_2}
    {D_i^j} = \frac{ n_r^i + mn_r^j}{(n_v^i + mn_v^j)\cdot(n_v^i + mn_v^j-1)}.
\end{equation}

The unknown $m$ is estimated based on the nodes in $u_j$ and the neighbors of $u_i$. Let $\eta_i$ be the number of neighboring nodes of $u_i$. Under a null model, we employ $\nicefrac{n_v^j}{\eta_i}$ to approximate the probability that $u_j$ appears among the neighboring vertices of $u_i$. The expected number of such vertices $m=\nicefrac{\eta_i}{n_v^j}$ is the reciprocal of the probability.

The above expectation is effective when $\eta_i>n_v^j$, i.e., $m\ge1$. This constraint arises from the fundamental property of $m$ that the number of mergeable candidates should be larger than 1 if $u_i$ has at least one connection outside itself. However, $\eta_i < n_v^j$ may occur. If $m < 1$, according to Eq.~\ref{eq4_2} where ${D_i^j} \propto \nicefrac{1}{m^2}$, the expected graph density ${D_i^j}$ will inflate quadratically as $m$ decreases. This effect is particularly severe when the connectivity within each vertex is relatively dense (i.e., small $\eta_i$ and large $n_v^j$), causing $m \ll 1$ and the merging condition to be overly strict. As a result, vertices from the same partition may fail to merge.
We therefore propose an alternative formulation in Eq.~\ref{eq4_2_m_1}. It replaces $m$ in the denominator of Eq.~\ref{eq4_2} with $m+1$, while $m$ in the numerator is retained to preserve the true expected edge connection.
In our experiments, we adopt this alternative formulation, and a comparison between Eq.~\ref{eq4_2} and Eq.~\ref{eq4_2_m_1} is provided in the ablation study.

\begin{equation}
    \label{eq4_2_m_1}
    {D_i^j} = \frac{n_r^i + mn_r^j}{(n_v^i + n_v^j + mn_v^j)\cdot(n_v^i + n_v^j + mn_v^j-1)}.
\end{equation}

In this null model, the expected number of common edges shared between $u_i$ and a single $u_j$ is $E[\zeta_i^j]={D_i^j}\cdot n_v^i \cdot n_v^j$. The threshold $\tau_i$ for merging is
\begin{equation}
    \label{eq4_3}
    \tau_i = \frac{\gamma_i \cdot E[\zeta_i^j]} {S_i} =\frac{\gamma_i \cdot {D_i^j}\cdot n_v^i \cdot n_v^j}{S_i}.
\end{equation}
where $\gamma_i = \min(\nicefrac{\zeta_i}{\zeta_j}, 1)$ is a balancing factor.

Strictly speaking, the numerator of $\tau_i$ represents the expected direct connections, which naturally serves as a lower bound for our connection similarity that captures both direct and indirect connections. However, the null model assumption may be invalid when two vertices exhibit significantly different sizes, as vertices under random connections should exhibit comparable sizes. Therefore, we introduce $\gamma$ in Eq.~\ref{eq4_3} to mitigate the impact of such size disparities between vertices. An ablation on this factor is also provided in the subsequent experiment.

\subsection{Two-stage graph partition}
\label{section:Tcd}

Edge weights in the HSG denote correlation strengths, allowing disconnected subgraphs to serve as initial partitions, as shown in Fig.~\ref{fig3_1}. Hence, graph partition is performed independently on each subgraph, which reduces the computational burden compared to performing it on the full graph.

This solution operates through an iterative two-stage process of search and merge. Initially, each node is treated as a singleton vertex. In the search stage, connection similarities are evaluated between neighboring vertex pairs. Any pair satisfying the merging criterion is assigned a common label. This continues until all vertices are examined. In the subsequent merge stage, vertices sharing the same label are integrated into a new vertex. A new graph is reconstructed with updated vertex set. The entire procedure repeats until no mergeable vertices remain, and the final vertex set is the optimal partitioning result.
A summary of the procedure is provided in Algorithm~\ref{alg:1}.

\begin{algorithm}[t]
    \caption{Connection Similarity based Graph Partition}
    \begin{small}
        \label{alg:1}
        \begin{algorithmic}[1]
            \REQUIRE $\text{A subgraph }G_k=\{u_i\}_{i=1}^{N_k}=\{(V_i, R_i)\}_{i=1}^{N_k}$
            \WHILE{True}
            \STATE $\text{Partitions } C\gets\{\}$, $flag \gets \text{False}$\\
            \COMMENT{Vertex initialization}
            \STATE $U_k \gets \text{RandomShuffleVertex}(G_k)$
            \FOR{$u_i$ in $U_k$}
            \STATE $\mathcal{N} \gets \textrm{Neighbor}(u_i, G_k)$
            \FOR{$u_j$ in $\mathcal{N}$}
            \STATE Calculate connection similarity $S_{i}^j$ and $S_j^{i}$ using Eq.~\ref{eq4_1}\\
            \COMMENT{Adaptive threshold calculation}
            \STATE Calculate graph density $D_i^j$ and $D_j^i$ using Eq.~\ref{eq4_2_m_1}\\
            \STATE Calculate threshold $\tau_i$ and $\tau_j$ using Eq.~\ref{eq4_3}\\
            \COMMENT{Merge decision}
            \IF{$S_i^j \geq \tau_i$ or $S_j^i \geq \tau_j$}
            \STATE $flag \gets \text{True}$
            \STATE $C(u_i) \gets \text{MergeAndLabel} (C(u_i), C(u_j))$
            \ENDIF
            \ENDFOR
            \ENDFOR
            \IF{$flag$ is False}
            \STATE break
            \ENDIF\\
            \COMMENT{Merge nodes and reconstruct graph}
            \STATE $G_k, U_k \gets \textrm{MergeAndReconstruct}(C)$
            \ENDWHILE
            \RETURN $C$
        \end{algorithmic}
    \end{small}
\end{algorithm}

\section{Experiments}
\label{section:exps}

This section evaluates the effectiveness of our method for event-based association.
We conduct experiments on eTram~\cite{vermaETraMEventbasedTraffic2024} and Ev-UAV~\cite{chenEventbasedTinyObject2025}. eTram is a novel event-based traffic perception dataset for MOT. It is curated using high spatial resolution ($1280\times 720$) event cameras across challenging lighting conditions, with object sizes ranging from approximately $8\times 8$ to $200\times 200$ pixels. Ev-UAV is the first event-based tiny object detection dataset, which contains tiny targets with an average size of $6.8\times 5.4$ pixels and provides event-wise labels. It covers challenging scenarios like high-brightness and low-light conditions.

On eTram, where the association scenario often resembles the image-based one-to-one matching, we compare our method against both event-based and image-based association methods under various detectors.
In contrast, the association on the Ev-UAV is many-to-many. We evaluate our method's ability to solve this underdetermined problem and to achieve robust tracking of tiny objects.
Finally, we conduct ablation studies and sensitivity experiments to assess the contribution of key components and the robustness of our method under various parameter configurations. Detailed parameter experiments and analyses are provided in the supplementary material.

\subsection{Evaluation metrics and model configurations}
\label{sec:metrc}

On eTram, we employ MT (Mostly Tracked), ML (Mostly Lost), MOTA~\cite{keniEvaluatingMultipleObject2008} (Multiple Object Tracking Accuracy), and IDF1 (Identification F-Score)~\cite{ristaniPerformanceMeasuresData2016b}. For completeness, we also report the detection F1 score which is related to MOTA.
On Ev-UAV, we still report the IDF1, while adopting the evaluation metrics from its original work~\cite{chenEventbasedTinyObject2025}, IoU and Acc (Accuracy). To assess tracking performance, we extend them to identity-aware versions by matching predicted identities with GT identities, referred to as Track IoU and Track Acc.

While our partition solution has no manually set hyperparameters, the fuzzy modeling pipeline has five parameters. They are the number of fitting samples $M$, hierarchical modeling levels $N$, correlation bias $\beta$, temporal scaling factor $s$ and temporal weight factor $\lambda$. We set $N=4$, $\beta=1$, and $s=0.001$. $M$ is not an important parameter, as it only affects the fitting accuracy of the quadratic B\'{e}zier curve, which only requires a minimum of three samples to fit. We set $M=7$ on eTram and $M=3$ on Ev-UAV since their batch periods are different.
The key temporal weight $\lambda$ is determined empirically. Specifically, $\lambda_{p1}=(1.001+\alpha\cdot(0.07-0.008\log_{10}w))^{\Delta t}$ for line-based positional errors, and $\lambda_{p2}=(1.001+(\alpha+1)\cdot{\tanh^2(0.02w)})^{\Delta t}$ for B\'{e}zier curve-based positional errors.

\subsection{Comparison methods}
\label{sec:metrc}
\textit{Event-based association methods.} To the best of our knowledge, existing event-based tracking methods operate in an online manner and lack dedicated multi-dimensional assignment methods. To demonstrate the performance improvement enabled by our method, we compare it against three prevailing event-based association methods: (1) Kalman Filter (KF) with Hungarian algorithm (HA)~\cite{barranco2018real, mentastiEventBasedObjectDetection2022,appsAsynchronousMultiObjectTracking2025, wangAsynchronousBlobTracker2024}, (2) neighbor searching (NS)~\cite{ruschTrackAERRealTimeEventBased2021, hu2022ecdt, alzugarayHASTEMultiHypothesisAsynchronous2020, hamannETAPEventbasedTracking2025}, and (3) IoU-based matching~\cite{chenAsynchronousTrackingDetectionAdaptive2019, chenEndtoEndLearningObject2020}. Besides these, we also compare with two event-based single object tracking (SOT) models, HDETrack~\cite{wangEventStreambasedVisual2024a} and TSETrack~\cite{zhangEfficientVisionTransformer2026}.

\textit{Image-based association methods.} Given the limited number of event-based association methods, we further compare our method with association modules from SOTA image-based MOT trackers.
It is important to note that these methods benefit from the one-to-one constraint and often utilize rich appearance features, while our method does not.
Despite these differences, we conduct a cross-modality comparison to thoroughly assess our tracking performance. This is done by substituting the association module in image-based trackers with our method. The comparative models are Permatrack~\cite{tokmakovLearningTrackObject2021}, UAVMOT~\cite{liuMultiObjectTrackingMeets2022}, OC-SORT~\cite{caoObservationCentricSORTRethinking2023}, LTTrack~\cite{linLTTrackRethinkingTracking2024}, TOPIC~\cite{caoTOPICParallelAssociation2025}, along with a multi-dimensional assignment based model LPT~\cite{liLearningGlobalObjective2022}. Since OC-SORT is a purely motion-based method that effectively combines and improves upon KF+HA and IoU-based matching, we adopt it as the primary baseline on Ev-UAV, rather than comparing against its individual components.
All models are evaluated under their standard training and testing protocols, and no extra post-processing (such as filtering) is applied to ensure a fair comparison.

\begin{table}
    \begin{center}
        \renewcommand\arraystretch{1.35}
        \begin{sloppypar}
            \setlength{\abovecaptionskip}{-0.15cm}
            \caption{MOT performance of different event-based association methods on eTram. The best values are highlighted in \textbf{bold}.}
            \vspace{0.2cm}
            \label{tab1}
            \begin{footnotesize}
                \begin{tabular}{
                    >{\centering}p{.9cm}|
                    >{\centering}p{0.4cm}|
                    >{\centering}p{.7cm}|
                    >{\centering}p{.6cm}|
                    >{\centering}p{.6cm}|
                    >{\centering}p{.6cm}|
                    >{\centering}p{.7cm}|
                    >{\centering\arraybackslash}p{.7cm}
                    }
                    \Xhline{1.2pt}
                    \multirow{2}*{Method}       & \multirow{2}*{Train} & Dataset & F1   & MT            & ML           & MOTA          & IDF1          \\
                                                &                      & Split   & (\%) & (\%)          & (\%)         & (\%)          & (\%)          \\
                    \Xhline{1.2pt}
                    \multirow{2}*{KF+HA}        & \multirow{2}*{No}    & Val     & 99.3 & 98.9          & \textbf{0.0} & 1.2           & 3.7           \\
                    \cline{3-8}
                                                &                      & Test    & 99.2 & 98.9          & 0.0          & 1.2           & 2.9           \\
                    \hline
                    \multirow{2}*{NS}           & \multirow{2}*{No}    & Val     & 99.3 & 99.4          & 0.1          & 57.7          & 30.9          \\
                    \cline{3-8}
                                                &                      & Test    & 99.2 & 98.9          & 0.0          & 65.0          & 38.8          \\
                    \hline
                    \multirow{2}*{IoU}          & \multirow{2}*{No}    & Val     & 99.3 & 99.4          & 0.1          & 87.2          & 30.7          \\
                    \cline{3-8}
                                                &                      & Test    & 99.2 & 98.9          & 0.0          & 90.6          & 22.8          \\
                    \hline
                    \multirow{2}*{\textbf{HSG}} & \multirow{2}*{No}    & Val     & 99.3 & \textbf{99.4} & 0.1          & \textbf{98.5} & \textbf{95.7} \\
                    \cline{3-8}
                                                &                      & Test    & 99.2 & \textbf{99.1} & \textbf{0.0} & \textbf{98.4} & \textbf{95.3} \\
                    \Xhline{1.2pt}
                \end{tabular}
            \end{footnotesize}
        \end{sloppypar}
    \end{center}
    \vspace{-0.2cm}
\end{table}

\subsection{Tracking on eTram}
\label{sec:etram}

\begin{figure}[!t]
    \begin{center}
        \centerline{\includegraphics[width=1\linewidth]{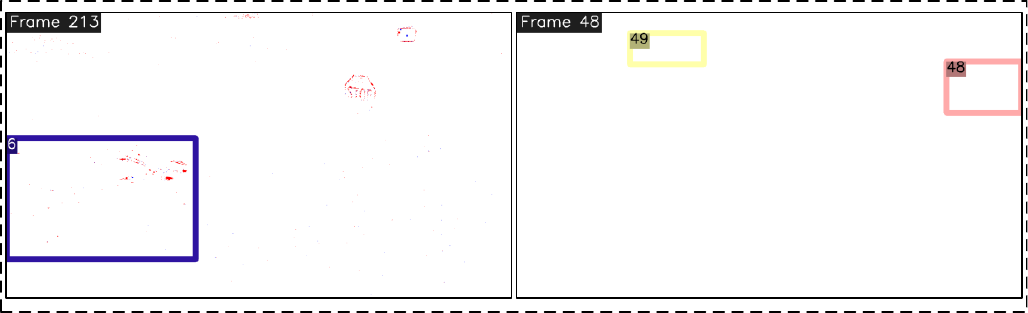}}
        \vspace{-0.2cm}
        \caption{Visualization of GT tracking annotations on eTram. Sequences (from left to right)are val\_night\_008 and val\_night\_011, showing GT boxes with few events that limit tracking performance during evaluation.}
        \label{fig5_2}
    \end{center}
    \vspace{-.3cm}
\end{figure}

\textit{Comparisons with event-based association methods.}
Experimental results of our method and event-based association methods under GT detections are presented in Table~\ref{tab1}.
Our method achieves the best tracking performance, attaining 98.5 MOTA and 95.7 IDF1 on the validation set and 98.4 MOTA and 95.3 IDF1 on the test set. It surpasses the second-best method by a significant margin (8\% in MOTA and 50\% in IDF1).
Though all methods are provided with GT detections, the reported F1 scores are not 1.0. We further analyze sequences with low F1. As visualized in Fig.~\ref{fig5_2}, some GT boxes contain very few or even zero events, making association infeasible. This leads to a reduction in true positives, which in turn lowers F1 and MOTA.

\textit{Comparisons with image-based association methods.}
We compare our method against SOTA image-based MOT association models and event-based SOT models under five different detectors, three image-based and two event-based.
As reported in Table~\ref{tab2}, our method consistently outperforms association baselines under three detectors, Permatrack, UAVMOT, and RVT. Notably, under the event-based detector RVT, our HSG surpasses OC-SORT with notable tracking performance improvements, especially in MT ($>$5\%), ML ($>$5\%) and IDF1 ($>$11\%).

When evaluated with the YOLOX detector~\cite{geYOLOXExceedingYOLO2021}, HSG shows slightly lower MOTA (0.6\%) and IDF1 (0.5\%) than OC-SORT on the validation dataset but achieves a clear lead ($>$3\%) in the other three metrics. We attribute OC-SORT's slight advantage in MOTA and IDF1 to OC-SORT's filtering of fragmented tracklets, which reduces false positives in detections and identity switches. Within the same YOLOX setting, HSG also outperforms TOPIC by over 10\% but falls slightly behind LTTrack (0.2\%-1.4\%).
When sharing the same detection results as TSETrack, our method consistently outperforms other methods such as OC-SORT and LPT, but shows a marginal performance drop of 0.9\% compared to TSETrack on the validation set. It is worth noting that the narrow margins against both LTTrack and TSETrack are within expectations. Given that our method operates without the one-to-one matching constraint, is completely training-free, and relies solely on kinematic cues, these minor performance drops represent a highly reasonable trade-off. Collectively, these results demonstrate the effectiveness of our approach in event-based association.

\begin{figure}[!t]
    \begin{center}
        \centerline{\includegraphics[width=1\linewidth]{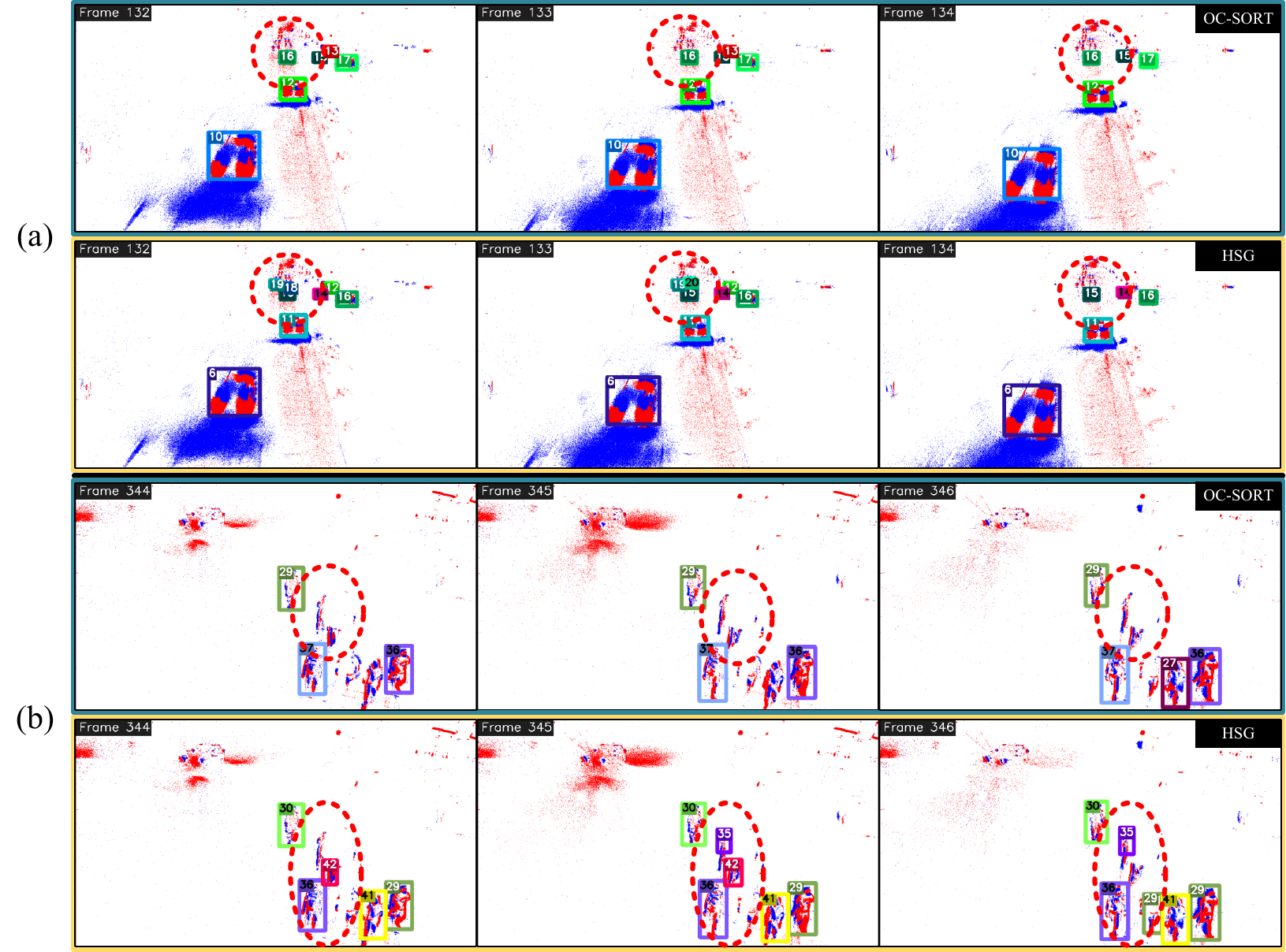}}
        \vspace{-0.2cm}
        \caption{Qualitative comparison of OC-SORT and our HSG on two hard eTram sequences under YOLOX detections. Shown case (a) represents small objects with missed detections, and (b) represents dense object distribution and occlusions. Red dashed ellipses highlight the challenging scenarios and result differences between two methods. Two sequences (from left to right) are val\_night\_001 and val\_night\_005.}
        \label{fig5_1}
    \end{center}
    \vspace{-.3cm}
\end{figure}

\begin{figure}[!t]
    \begin{center}
        \centerline{\includegraphics[width=1\linewidth]{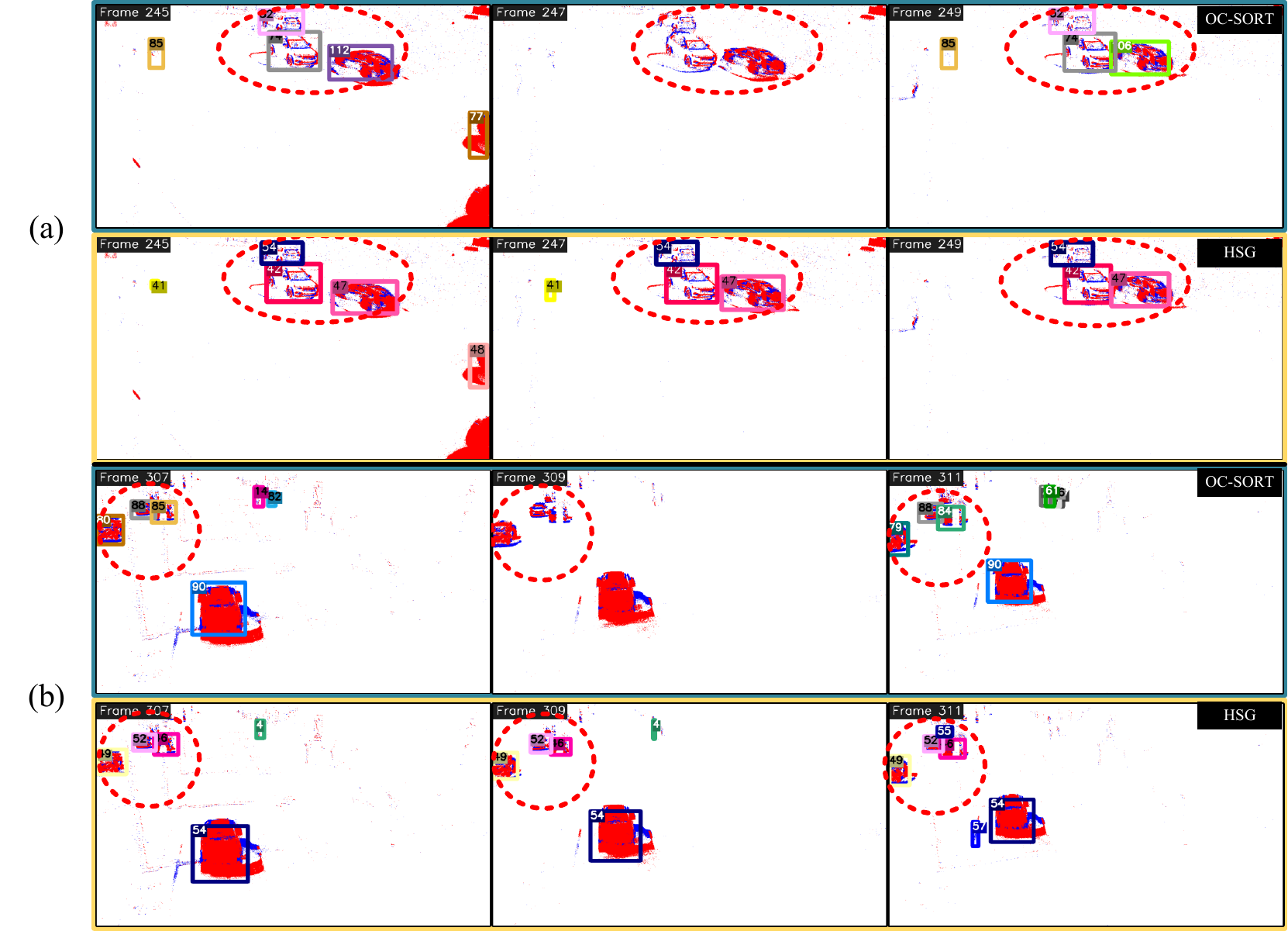}}
        \vspace{-0.2cm}
        \caption{Qualitative comparison of OC-SORT and our HSG on two hard eTram sequences under RVT detections. Shown case (a) represents objects with partial occlusions, and (b) represents dense object distribution and occlusions. Red dashed ellipses highlight the challenging scenarios and result differences between two methods. Two sequences (from left to right) are test\_day\_005 and test\_day\_006.}
        \label{fig5_1_5}
    \end{center}
    \vspace{-.3cm}
\end{figure}

\textit{Case Study}. Fig.~\ref{fig5_1} and Fig.~\ref{fig5_1_5} visualize several challenging tracking scenarios comparing baseline OC-SORT and our HSG under YOLOX and RVT detections, respectively. In Fig.~\ref{fig5_1}, two representative cases are shown, (a) small objects with missed detections, and (b) missed detections in dense distributions. Tracking differences between OC-SORT and our method are highlighted in red dashed ellipses. In case (a), some detections of small objects are temporarily missing in frames 132 and 133. In case (b), some pedestrian detections are also missing in frames 344 and 346. Our method successfully associates fragmented tracklets in both scenarios, while OC-SORT fails and drops them due to insufficient tracking hits over time.

\begin{table*}
    \begin{center}
        \renewcommand\arraystretch{1.25}
        \begin{sloppypar}
            \setlength{\abovecaptionskip}{-0.15cm}
            \caption{Tracking Performance: Our Method vs. SOTA Image-based Trackers under multiple detectors on eTram. Methods sharing the same detection results are grouped in identically colored blocks. In tracking type, -L means linear assignment-based and -M means multi-dimensional assignment-based. In association feature type, M- means motion-based, A- means appearance-based, and Post means post-processing. `Train' column indicates whether the association method utilizes additional cues beyond bounding-box coordinates from a trained network. Regarding performance comparisons, \textcolor{purple}{Purple} indicates the positive performance gap of our method compared to the \underline{baseline}, while \textcolor{DarkGreen}{Dark Green} indicates a negative performance gap.}
            \vspace{0.3cm}
            \label{tab2}
            \begin{footnotesize}
                \begin{tabular}{
                    >{\centering}p{1.1cm}|
                    >{\centering}p{.9cm}|
                    >{\centering}p{1.4cm}|
                    >{\centering}p{2.3cm}|
                    >{\centering}p{0.5cm}|
                    >{\centering}p{1.4cm}|
                    >{\centering}p{.5cm}|
                    >{\centering}p{.7cm}|
                    >{\centering}p{.6cm}|
                    >{\centering}p{.7cm}|
                    >{\centering}p{.8cm}|
                    >{\centering}p{.8cm}|
                    >{\centering\arraybackslash}p{.8cm}
                    }
                    \Xhline{1.2pt}
                    \multirow{2}*{Method}                          & Tracking              & Association                & Association                                                            & \multirow{2}*{Train}  & One-to-one            & \multirow{2}*{Year}  & Dataset               & F1                                   & MT                                      & ML                                   & MOTA                                   & IDF1                                   \\

                                                                   & type                  & feature type               & techniques                                                             &                       & constraint            &                      & Split                 & (\%)                                 & (\%)                                    & (\%)                                 & (\%)                                   & (\%)                                   \\
                    \Xhline{1.2pt}
                    \rowcolor{Melon!60}                            &                       &                            &                                                                        &                       &                       &                      & Val                   & \underline{54.7}                     & \underline{30.6}                        & \underline{18.0}                     & \underline{16.5}                       & \underline{49.0}                       \\
                    \rowcolor{Melon!60} \multirow{-2}*{Permatrack} & \multirow{-2}*{MOT-L} & \multirow{-2}*{M-+A-}      & \multirow{-2}*{HA+ConvGRU~\cite{ballasDelvingDeeperConvolutional2016}} & \multirow{-2}* {Yes}  & \multirow{-2}* {Yes}  & \multirow{-2}*{2021} & Test                  & \underline{47.9}                     & \underline{13.4}                        & \underline{21.5}                     & \underline{17.1}                       & \underline{41.6}                       \\
                    \hline
                    \rowcolor{Melon!60}
                                                                   &                       &                            &                                                                        &                       &                       &                      &                       & 54.7                                 & 30.6                                    & 18.0                                 & 16.9                                   & 50.4                                   \\
                    \rowcolor{Melon!60}
                                                                   &                       &                            &                                                                        &                       &                       &                      & \multirow{-2}* {Val}  & \textcolor{purple}{\textbf{(0.0)}}   & \textcolor{purple}{\textbf{(0.0)}}      & \textcolor{purple}{\textbf{(0.0)}}   & \textcolor{purple}{\textbf{(+0.4)}}    & \textcolor{purple}{\textbf{(+1.4)}}    \\
                    \rowcolor{Melon!60}
                                                                   &                       &                            &                                                                        &                       &                       &                      &                       & 47.9                                 & 13.4                                    & 21.5                                 & 17.6                                   & 43.7                                   \\
                    \rowcolor{Melon!60}
                    \multirow{-4}*{HSG}                            & \multirow{-4}*{MOT-M} & \multirow{-4}*{M-}         & \multirow{-4}{*}{Graph-based optim.}                                   & \multirow{-4}* {{No}} & \multirow{-4}* {{No}} & \multirow{-4}* {--}  & \multirow{-2}* {Test} & \textcolor{purple}{\textbf{(0.0)}}   & \textcolor{purple}{\textbf{(0.0)}}      & \textcolor{purple}{\textbf{(0.0)}}   & \textcolor{purple}{\textbf{(+0.5)}}    & \textcolor{purple}{\textbf{(+2.1)}}    \\
                    \Xhline{1.2pt}
                    \rowcolor{Peach!38}
                                                                   &                       &                            &                                                                        &                       &                       &                      & Val                   & \underline{65.6}                     & \underline{73.0}                        & \underline{2.5}                      & \underline{19.8}                       & \underline{57.7}                       \\
                    \noalign{\vskip-\arrayrulewidth}
                    \rowcolor{Peach!38}
                    \multirow{-2}*{UAVMOT}                         & \multirow{-2}*{MOT-L} & \multirow{-2}*{M-+A-}      & \multirow{-2}*{KF+HA+IoU+NS}                                           & \multirow{-2}* {Yes}  & \multirow{-2}* {Yes}  & \multirow{-2}*{2022} & Test                  & \underline{68.6}                     & \underline{51.9}                        & \underline{4.3}                      & \underline{34.1}                       & \underline{59.4}                       \\
                    \hline
                    \rowcolor{Peach!38}
                                                                   &                       &                            &                                                                        &                       &                       &                      &                       & 65.9                                 & 73.8                                    & 2.4                                  & 21.5                                   & 59.1                                   \\
                    \rowcolor{Peach!38}
                                                                   &                       &                            &                                                                        &                       &                       &                      & \multirow{-2}* {Val}  & \textcolor{purple}{\textbf{(+0.3)}}  & \textcolor{purple}{\textbf{(+0.8)}}     & \textcolor{purple}{\textbf{(-0.1)}}  & \textcolor{purple}{\textbf{(+1.7)}}    & \textcolor{purple}{\textbf{(+1.4)}}    \\
                    \rowcolor{Peach!38}
                                                                   &                       &                            &                                                                        &                       &                       &                      &                       & 68.9                                 & 52.9                                    & 4.3                                  & 36.3                                   & 60.7                                   \\
                    \rowcolor{Peach!38}
                    \multirow{-4}*{HSG}                            & \multirow{-4}*{MOT-L} & \multirow{-4}*{M-}         & \multirow{-4}{*}{--}                                                   & \multirow{-4}* {{No}} & \multirow{-4}* {{No}} & \multirow{-4}* {--}  & \multirow{-2}* {Test} & \textcolor{purple}{\textbf{(+0.3)}}  & \textcolor{purple}{\textbf{(+1.0)}}     & \textcolor{purple}{\textbf{(0.0)}}   & \textcolor{purple}{\textbf{(+2.2)}}    & \textcolor{purple}{\textbf{(+1.3)}}    \\
                    \Xhline{1.2pt}
                    \rowcolor{Dandelion!30}
                                                                   &                       &                            &                                                                        &                       &                       &                      & Val                   & 57.2                                 & --                                      & --                                   & --                                     & --                                     \\
                    \rowcolor{Dandelion!30}
                    \multirow{-2}*{YOLOX-x}                        & \multirow{-2}*{--}    & \multirow{-2}*{--}         & \multirow{-2}*{--}                                                     & \multirow{-2}* {--}   & \multirow{-2}* {--}   & \multirow{-2}*{2023} & Test                  & 51.1                                 & --                                      & --                                   & --                                     & --                                     \\
                    \hline
                    \rowcolor{Dandelion!30}
                                                                   &                       &                            &                                                                        &                       &                       &                      & Val                   & \underline{55.1}                     & \underline{31.2}                        & \underline{28.2}                     & \underline{27.9}                       & \underline{52.9}                       \\
                    \rowcolor{Dandelion!30}
                    \multirow{-2}*{OC-SORT}                        & \multirow{-2}*{MOT-L} & \multirow{-2}*{M-}         & \multirow{-2}*{KF+HA+IoU}                                              & \multirow{-2}* {Yes}  & \multirow{-2}* {Yes}  & \multirow{-2}*{2023} & Test                  & \underline{46.5}                     & \underline{9.2}                         & \underline{45.3}                     & \underline{25.4}                       & \underline{44.8}                       \\
                    \hline
                    \rowcolor{Dandelion!30}
                                                                   &                       &                            &                                                                        &                       &                       &                      & Val                   & {58.7}                               & {44.6}                                  & {23.4}                               & {27.5}                                 & {53.8}                                 \\
                    \rowcolor{Dandelion!30}
                                                                   &                       & \multirow{-2}*{M-+A-}      & \multirow{-2}*{HA+IoU+NS+}                                             &                       &                       &                      & Test                  & {53.0}                               & {15.4}                                  & {39.1}                               & {29.3}                                 & {47.6}                                 \\
                    \Xcline{3-3}{1.3pt}\Xcline{8-13}{1.3pt}
                    \rowcolor{Dandelion!30}
                                                                   &                       &                            &                                                                        &                       &                       &                      & Val                   & 59.2                                 & 47.7                                    & 23.1                                 & 26.3                                   & 54.2                                   \\
                    \rowcolor{Dandelion!30}
                    \multirow{-4}*{LTTrack}                        & \multirow{-4}*{MOT-L} & \multirow{-2}*{M-+A-+Post} & \multirow{-2}*{FastReID~\cite{he2020fastreid}}                         & \multirow{-4}*{Yes}   & \multirow{-4}* {Yes}  & \multirow{-4}*{2024} & Test                  & 55.0                                 & 18.6                                    & 38.7                                 & 30.5                                   & 49.3                                   \\
                    \hline
                    \rowcolor{Dandelion!30}
                                                                   &                       &                            &                                                                        &                       &                       &                      & Val                   & 48.2                                 & 10.5                                    & 28.6                                 & 17.4                                   & 20.8                                   \\
                    \rowcolor{Dandelion!30}
                                                                   &                       & \multirow{-2}*{M-+A-}      &                                                                        &                       &                       &                      & Test                  & 37.7                                 & 1.9                                     & 48.8                                 & 13.5                                   & 15.7                                   \\
                    \Xcline{3-3}{1.3pt}\Xcline{8-13}{1.3pt}
                    \rowcolor{Dandelion!30}
                                                                   &                       &                            &                                                                        &                       &                       &                      & Val                   & 48.9                                 & 21.7                                    & 26.8                                 & 8.3                                    & 22.2                                   \\
                    \rowcolor{Dandelion!30}
                    \multirow{-4}*{TOPIC}                          & \multirow{-4}*{MOT-L} & \multirow{-2}*{M-+A-+Post} & \multirow{-4}*{OC-SORT+FastReID}                                       & \multirow{-4}*{Yes}   & \multirow{-4}* {Yes}  & \multirow{-4}*{2025} & Test                  & 40.2                                 & 4.9                                     & 45.2                                 & 9.2                                    & 18.0                                   \\
                    \hline
                    \rowcolor{Dandelion!30}
                                                                   &                       &                            &                                                                        &                       &                       &                      & Val                   & 56.8                                 & 44.8                                    & 22.8                                 & 19.5                                   & 36.5                                   \\
                    \rowcolor{Dandelion!30}
                    \multirow{-2}*{LPT}                            & \multirow{-2}*{MOT-M} & \multirow{-2}*{M-+A-}      & \multirow{-2}*{Graph-based optim.}                                     & \multirow{-2}* {Yes}  & \multirow{-2}* {Yes}  & \multirow{-2}*{2022} & Test                  & 52.3                                 & 16.5                                    & 39.4                                 & 24.8                                   & 34.2                                   \\
                    \hline
                    \rowcolor{Dandelion!30}
                                                                   &                       &                            &                                                                        &                       &                       &                      &                       & 58.1                                 & 42.3                                    & 23.2                                 & 27.3                                   & 52.6                                   \\
                    \rowcolor{Dandelion!30}
                                                                   &                       &                            &                                                                        &                       &                       &                      & \multirow{-2}* {Val}  & \textcolor{purple}{\textbf{(+3.0)}}  & \textcolor{purple}{\textbf{(+11.1)}}    & \textcolor{purple}{\textbf{(-5.0)}}  & \textcolor{DarkGreen}{\textbf{(-0.6)}} & \textcolor{DarkGreen}{\textbf{(-0.3)}} \\
                    \rowcolor{Dandelion!30}
                                                                   &                       &                            &                                                                        &                       &                       &                      &                       & 52.1                                 & 16.7                                    & 39.4                                 & 29.0                                   & 47.2                                   \\
                    \rowcolor{Dandelion!30}
                    \multirow{-4}*{HSG}                            & \multirow{-4}*{MOT-M} & \multirow{-4}*{M-}         & \multirow{-4}{*}{Graph-based optim.}                                   & \multirow{-4}* {{No}} & \multirow{-4}* {{No}} & \multirow{-4}* {--}  & \multirow{-2}* {Test} & \textcolor{purple} {\textbf{(+5.6)}} & \textcolor{purple} {\textbf{(+7.5)}}    & \textcolor{purple}{\textbf{(-5.9)}}  & \textcolor{purple}{\textbf{(+3.6)}}    & \textcolor{purple}{\textbf{(+2.4)}}    \\
                    \Xhline{1.2pt}
                    \rowcolor{Dandelion!10}
                                                                   &                       &                            &                                                                        &                       &                       &                      & Val                   & 40.7                                 & --                                      & --                                   & --                                     & --                                     \\
                    \rowcolor{Dandelion!10}
                    \multirow{-2}*{RVT-b}                          & \multirow{-2}*{--}    & \multirow{-2}*{--}         & \multirow{-2}*{--}                                                     & \multirow{-2}* {--}   & \multirow{-2}* {--}   & \multirow{-2}*{2023} & Test                  & 39.1                                 & --                                      & --                                   & --                                     & --                                     \\
                    \hline
                    \rowcolor{Dandelion!10}
                                                                   &                       &                            &                                                                        &                       &                       &                      & Val                   & \underline{43.7}                     & \underline{10.8}                        & \underline{18.8}                     & \underline{-11.7}                      & \underline{29.4}                       \\
                    \rowcolor{Dandelion!10}
                    \multirow{-2}*{OC-SORT}                        & \multirow{-2}*{MOT-L} & \multirow{-2}*{M-}         & \multirow{-2}*{KF+HA+IoU}                                              & \multirow{-2}* {Yes}  & \multirow{-2}* {Yes}  & \multirow{-2}*{2023} & Test                  & \underline{43.5}                     & \underline{7.2}                         & \underline{26.5}                     & \underline{1.7}                        & \underline{28.0}                       \\
                    \hline
                    \rowcolor{Dandelion!10}
                                                                   &                       &                            &                                                                        &                       &                       &                      & Val                   & 36.9                                 & 84.4                                    & 1.5                                  & -178.5                                 & 18.8                                   \\
                    \rowcolor{Dandelion!10}
                    \multirow{-2}*{LPT}                            & \multirow{-2}*{MOT-M} & \multirow{-2}*{M-+A-}      & \multirow{-2}*{Graph-based optim.}                                     & \multirow{-2}* {Yes}  & \multirow{-2}* {Yes}  & \multirow{-2}*{2022} & Test                  & 36.3                                 & 57.1                                    & 3.7                                  & -146.3                                 & 17.1                                   \\
                    \hline
                    \rowcolor{Dandelion!10}
                                                                   &                       &                            &                                                                        &                       &                       &                      &                       & 45.9                                 & 15.8                                    & 13.1                                 & -10.4                                  & 40.9                                   \\
                    \rowcolor{Dandelion!10}
                                                                   &                       &                            &                                                                        &                       &                       &                      & \multirow{-2}* {Val}  & \textcolor{purple}{\textbf{(+2.2)}}  & \textcolor{purple}{\textbf{(+5.0)}}     & \textcolor{purple}{\textbf{(-5.7)}}  & \textcolor{purple}{\textbf{(+1.3)}}    & \textcolor{purple}{\textbf{(+11.5)}}   \\
                    \rowcolor{Dandelion!10}
                                                                   &                       &                            &                                                                        &                       &                       &                      &                       & 46.6                                 & 15.5                                    & 16.1                                 & 2.1                                    & 40.4                                   \\
                    \rowcolor{Dandelion!10}
                    \multirow{-4}*{HSG}                            & \multirow{-4}*{MOT-M} & \multirow{-4}*{M-}         & \multirow{-4}{*}{Graph-based optim.}                                   & \multirow{-4}* {{No}} & \multirow{-4}* {{No}} & \multirow{-4}* {--}  & \multirow{-2}* {Test} & \textcolor{purple} {\textbf{(+3.1)}} & \textcolor{purple} {\textbf{(+8.3)}}    & \textcolor{purple}{\textbf{(-10.4)}} & \textcolor{purple}{\textbf{(+0.4)}}    & \textcolor{purple}{\textbf{(+12.4)}}   \\
                    \Xhline{1.2pt}
                                                                   &                       &                            &                                                                        &                       &                       &                      & Val                   & {68.7}                               & {69.9}                                  & {3.0}                                & {36.9}                                 & {66.5}                                 \\
                    \multirow{-2}*{HDETrack}                       & \multirow{-2}*{SOT}   & \multirow{-2}*{M-+A-}      & \multirow{-2}*{E2E Transformer}                                        & \multirow{-2}* {Yes}  & \multirow{-2}* {Yes}  & \multirow{-2}*{2024} & Test                  & {59.2}                               & {48.8}                                  & {7.2}                                & {17.8}                                 & {56.0}                                 \\
                    \hline
                    \rowcolor{Dandelion!5}

                                                                   &                       &                            &                                                                        &                       &                       &                      & Val                   & \underline{74.5}                     & \underline{83.3}                        & \underline{3.0}                      & \underline{48.2}                       & \underline{72.3}                       \\
                    \rowcolor{Dandelion!5}
                    \multirow{-2}*{TSETrack}                       & \multirow{-2}*{SOT}   & \multirow{-2}*{M-+A-}      & \multirow{-2}*{E2E Transformer}                                        & \multirow{-2}* {Yes}  & \multirow{-2}* {Yes}  & \multirow{-2}*{2026} & Test                  & \underline{65.4}                     & \underline{61.3}                        & \underline{7.0}                      & \underline{29.9}                       & \underline{62.1}                       \\
                    \hline
                    \rowcolor{Dandelion!5}
                                                                   &                       &                            &                                                                        &                       &                       &                      & Val                   & 72.8                                 & 76.6                                    & 4.0                                  & 47.3                                   & 67.0                                   \\
                    \rowcolor{Dandelion!5}
                    \multirow{-2}*{OC-SORT}                        & \multirow{-2}*{MOT-L} & \multirow{-2}*{M-}         & \multirow{-2}*{KF+HA+IoU}                                              & \multirow{-2}* {Yes}  & \multirow{-2}* {Yes}  & \multirow{-2}*{2023} & Test                  & 63.9                                 & 52.3                                    & 9.2                                  & 30.8                                   & 56.5                                   \\
                    \hline
                    \rowcolor{Dandelion!5}
                                                                   &                       &                            &                                                                        &                       &                       &                      & Val                   & 72.1                                 & 82.6                                    & 3.3                                  & 41.3                                   & 53.0                                   \\
                    \rowcolor{Dandelion!5}
                    \multirow{-2}*{LPT}                            & \multirow{-2}*{MOT-M} & \multirow{-2}*{M-+A-}      & \multirow{-2}*{Graph-based optim.}                                     & \multirow{-2}* {Yes}  & \multirow{-2}* {Yes}  & \multirow{-2}*{2022} & Test                  & 64.1                                 & 39.9                                    & 12.2                                 & 37.2                                   & 39.8                                   \\
                    \hline
                    \rowcolor{Dandelion!5}

                                                                   &                       &                            &                                                                        &                       &                       &                      &                       & 75.0                                 & 83.2                                    & 3.0                                  & 49.9                                   & 71.4                                   \\
                    \rowcolor{Dandelion!5}
                                                                   &                       &                            &                                                                        &                       &                       &                      & \multirow{-2}* {Val}  & \textcolor{purple}{\textbf{(+0.5)}}  & \textcolor{DarkGreen}{\textbf{(-0.1)}}  & \textcolor{purple}{\textbf{(0.0)}}   & \textcolor{purple}{\textbf{(+1.7)}}    & \textcolor{DarkGreen}{\textbf{(-0.9)}} \\
                    \rowcolor{Dandelion!5}
                                                                   &                       &                            &                                                                        &                       &                       &                      &                       & 66.3                                 & 61.1                                    & 7.0                                  & 32.9                                   & 63.6                                   \\
                    \rowcolor{Dandelion!5}
                    \multirow{-4}*{{HSG}}                          & \multirow{-4}*{MOT-M} & \multirow{-4}*{M-}         & \multirow{-4}{*}{Graph-based optim.}                                   & \multirow{-4}* {{No}} & \multirow{-4}* {{No}} & \multirow{-4}* {--}  & \multirow{-2}* {Test} & \textcolor{purple} {\textbf{(+0.9)}} & \textcolor{DarkGreen} {\textbf{(-0.2)}} & \textcolor{purple}{\textbf{(0.0)}}   & \textcolor{purple}{\textbf{(+3.0)}}    & \textcolor{purple}{\textbf{(+1.5)}}    \\
                    \Xhline{1.2pt}
                \end{tabular}
            \end{footnotesize}
        \end{sloppypar}
    \end{center}
\end{table*}

In Fig.~\ref{fig5_1_5}, we also present two cases, (a) objects partially occluded by background, and (b) dense distribution with object occlusions. In these cases, our method still successfully associates the tracklets, while OC-SORT fails. These results demonstrate that our method achieves higher tracking accuracy than a strong motion-based baseline. Additionally, we observe that in case (a), our method fails to associate a tracklet on the left side of frame 249 where OC-SORT succeeds. This is because the corresponding detection bounding box contains nearly zero events, making our event-based association infeasible. A similar issue appears in frame 311 of case (b) where OC-SORT produces identities for three overlapping detections.

\subsection{Tracking on Ev-UAV}

\begin{figure}[!t]
    \begin{center}
        \centerline{\includegraphics[width=1\linewidth]{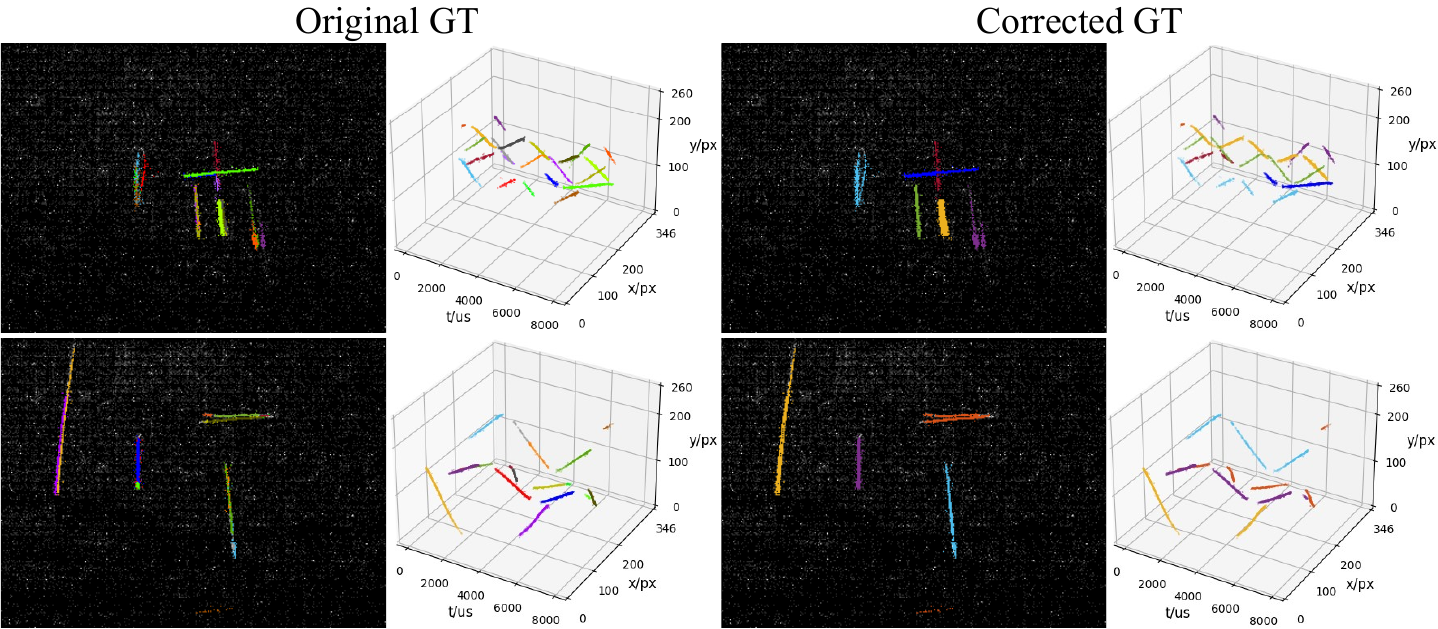}}
        \vspace{-0.2cm}
        \caption{Visualization of the GT tracking labels and our corrected tracking labels on Ev-UAV. Sequences (from top to bottom): val\_16 and test\_9.}
        \label{fig5_3}
    \end{center}
    \vspace{-.3cm}
\end{figure}

\begin{figure*}[!t]
    \begin{center}
        \centerline{\includegraphics[width=1\linewidth]{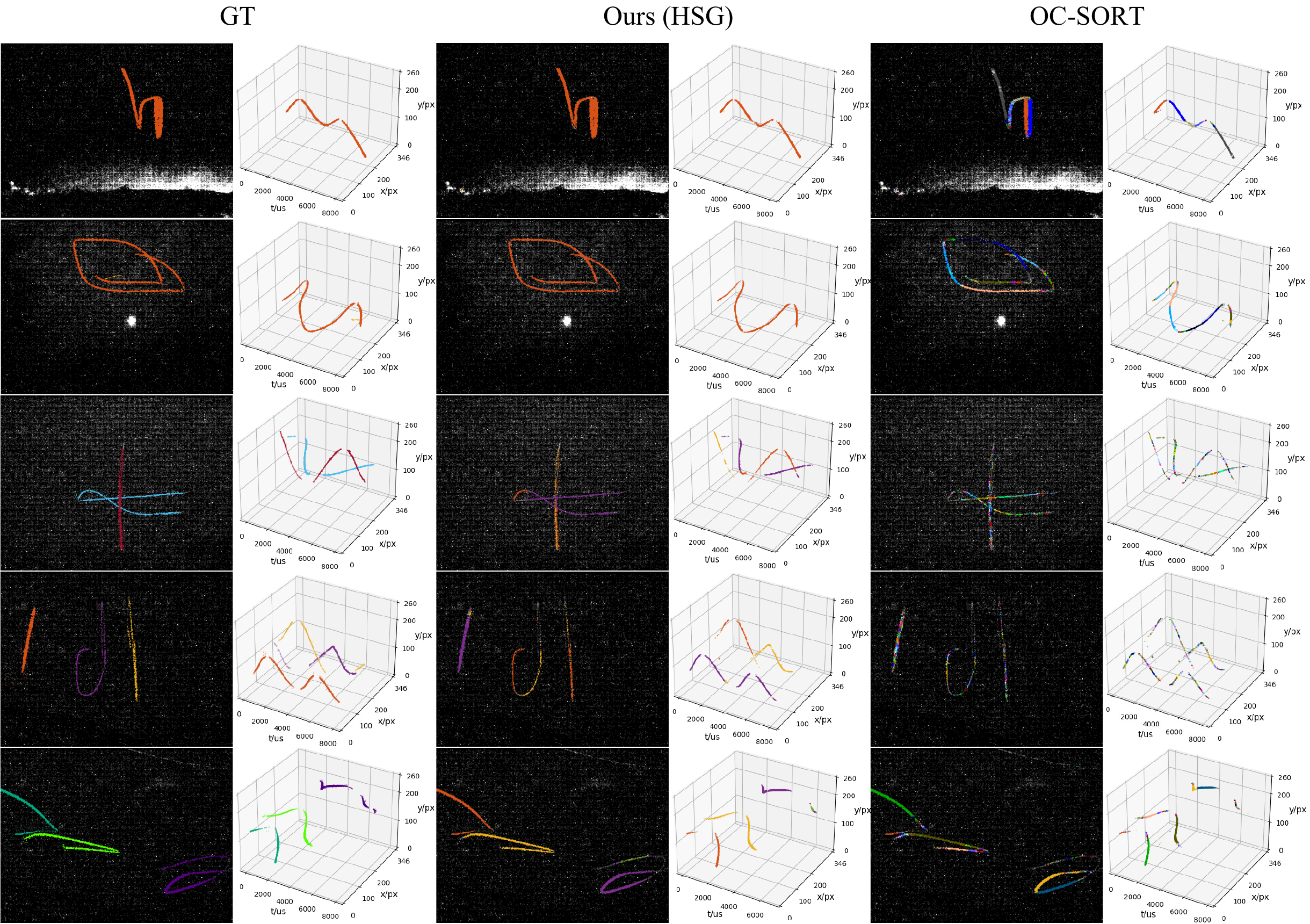}}
        \vspace{-0.2cm}
        \caption{Visualization of the object trajectories of GT, proposed HSG, and OC-SORT on Ev-UAV. Sequences feature the challenge of associating tracklets under large spatio-temporal gaps and multiple motion patterns. Sequences (from top to bottom): val\_11$\#$, val\_20$\#$, test\_6$*$, test\_10$*$, and test\_13$*$. $\#$ denotes the unmodified sequence, while $*$ denotes the corrected sequence.}
        \label{fig5_4}
    \end{center}
    \vspace{-.3cm}
\end{figure*}

As Ev-UAV is a detection benchmark, the originally provided tracking labels contain some inaccuracies, we corrected and refined the tracking labels on test and validation sets. This was done by manually comparing UAV motion patterns in 2D plane and 3D space to ensure reliable tracking annotations, as illustrated in Fig.~\ref{fig5_3}. More tracking label correction details please refer to the supplementary material, and the corrected dataset will be published in the future~\footnote{\url{https://github.com/zhazhabiu/HSG-Event-based-Association}}. The total number of sequences in test and validation sets is 48. Among these, 20 sequences were refined, including 14 in the test set and 6 in the validation set. To ensure transparency, we report evaluation results both on the remaining 28 unmodified sequences and on the full set with corrected sequences in Table~\ref{tab3}. The rectified dataset with updated tracking annotations will be made publicly available in our repository.

\begin{table}
    \begin{center}
        \renewcommand\arraystretch{1.35}
        \begin{sloppypar}
            \setlength{\abovecaptionskip}{-0.15cm}
            \vspace{-0.4cm}
            \caption{Tracking performance on Ev-UAV. $*$ denotes the full set with corrected tracking annotations. }
            \vspace{.1cm}
            \label{tab3}
            \begin{tabular}{
                >{\centering}p{1.3cm}|
                >{\centering}p{.8cm}|
                >{\centering}p{1.5cm}|
                >{\centering}p{1.5cm}|
                >{\centering\arraybackslash}p{1.5cm}
                }
                \Xhline{1.2pt}
                \multirow{2}*{Method}  & \multirow{2}*{Dataset} & Track IoU                                   & Track Acc                                   & IDF1                                        \\
                                       &                        & (\%)                                        & (\%)                                        & (\%)                                        \\
                \Xhline{1.2pt}
                \multirow{2}*{OC-SORT} & Val                    & 21.2                                        & 13.1                                        & 9.6                                         \\
                \cline{2-5}
                                       & Test                   & 23.3                                        & 16.5                                        & 8.8                                         \\
                \hline
                \multirow{2}*{{HSG}}   & Val                    & 51.1 \textcolor{purple}{\textbf{(+29.9)}}   & 62.3 \textcolor{purple}{\textbf{(+49.2)}}   & 66.2 \textcolor{purple}{\textbf{(+56.6)}}   \\
                \cline{2-5}
                                       & Test                   & {47.8} \textcolor{purple}{\textbf{(+24.5)}} & {49.6} \textcolor{purple}{\textbf{(+33.1)}} & {57.3} \textcolor{purple}{\textbf{(+48.5)}} \\
                \Xhline{1.2pt}
                \multirow{2}*{OC-SORT} & Val$*$                 & 15.3                                        & 11.7                                        & 4.7                                         \\
                \cline{2-5}
                                       & Test$*$                & 15.3                                        & 13.3                                        & 3.9                                         \\
                \hline
                \multirow{2}*{{HSG}}   & Val$*$                 & {49.9} \textcolor{purple}{\textbf{(+34.6)}} & {63.7} \textcolor{purple}{\textbf{(+52.0)}} & {81.3} \textcolor{purple}{\textbf{(+76.6)}} \\
                \cline{2-5}
                                       & Test$*$                & {47.2} \textcolor{purple}{\textbf{(+31.9)}} & {52.7} \textcolor{purple}{\textbf{(+39.4)}} & {74.0} \textcolor{purple}{\textbf{(+70.1)}} \\
                \Xhline{1.2pt}
            \end{tabular}
        \end{sloppypar}
    \end{center}
    \vspace{-.2cm}
\end{table}

The upstream detector is EV-SpSegNet~\cite{chenEventbasedTinyObject2025}, a segmentation network that extracts object events from raw event data. Since it performs binary segmentation, we apply a region growing algorithm with a searching radius of $r=7$ pixels within every 300 ms interval to cluster the segmented events into tracklets. These tracklets are then fed into the association pipeline.

As shown in Table~\ref{tab3}, our approach outperforms OC-SORT by over 2$\times$ across all metrics on both original and corrected Ev-UAV. Fig.~\ref{fig5_4} shows several challenging sequences containing tracklets with large spatio-temporal gaps and multiple motion patterns. Additional experiments with varying region growing radius $r$ are provided in the supplementary material. All these results demonstrate that our method successfully associates long-range fragmented tracklets into complete trajectories.

The Ev-UAV dataset presents a classic and challenging tiny object tracking scenario. Different from image-based small object tracking, events here offer no appearance cues and the association problem is further complicated by its many-to-many nature. Despite these challenges, our method demonstrates robust and accurate associations, reliably linking isolated tracklets and effectively addressing the inherent difficulties of tiny object tracking.

\subsection{Computational Efficiency}
\label{section:CE}
Although multi-dimensional assignment sacrifices some computational efficiency for accuracy gains, we rigorously analyze the time complexity of our proposed method. Our method consists of two stages, graph modeling and graph partitioning. Let $\mathcal{N}$ denote the number of tracklets to be associated, and $\zeta$ ($\zeta \ll \mathcal{N}$) denote the candidates of each tracklet after initial sparsification. In the modeling stage, metric calculations operate in $O(\mathcal{N}^2)$, and K-means with $I$ iterations and $N+1$ centers operates in $O(\mathcal{N} \cdot I \cdot \zeta \cdot (N+1))$. Consequently, the whole time complexity of modeling is $O(\mathcal{N}\cdot\zeta + \mathcal{N}^2)$, which is dominated by $O(\mathcal{N}^2)$ since our implementation is in serial. After modeling, the HSG consists of $\mathcal{N}$ nodes and $E\approx \nicefrac{(\mathcal{N}\cdot \zeta)}{2}$ edges. In the first partitioning iteration, computing the connectivity similarity for all vertices costs $O(\mathcal{N}\cdot \zeta^2)$. Since merging costs $O(E)$ which is negligible compared to $O(\mathcal{N}\cdot \zeta^2)$, the time complexity of a single search-merge is $O(\mathcal{N}\cdot \zeta^2)$. Each iteration reduces the number of vertices by at least half, so partitioning requires at most $O(\log_2\mathcal{N})$ iterations. The time complexity of the whole graph partitioning is $O(\mathcal{N}\cdot \zeta^2\cdot \log_2\mathcal{N})$. Therefore, the overall time complexity of our method is $O(\mathcal{N}^2 + \mathcal{N}\cdot \zeta^2\cdot \log_2\mathcal{N})$.

\begin{table}
    \begin{center}
        \renewcommand\arraystretch{1.35}
        \begin{sloppypar}
            \vspace{-.2cm}
            \setlength{\abovecaptionskip}{-0.15cm}
            \caption{Running efficiency comparing with association baselines under YOLOX detections on eTram. The best values are in ``\textbf{bold}''.}
            \vspace{0.1cm}
            \label{tab4_2}
            \begin{tabular}{
                >{\centering}p{1.3cm}|
                >{\centering}p{.9cm}|
                >{\centering}p{1.0cm}|
                >{\centering}p{1.3cm}|
                >{\centering\arraybackslash}p{1.3cm}
                }
                \Xhline{1.2pt}
                Method                 & Dataset & FPS             & EPS($\times10^6$) & Hardware           \\
                \Xhline{1.2pt}
                \multirow{2}*{LTTrack} & Val     & 3.31            & --                & \multirow{6}*{GPU} \\
                \cline{2-4}
                                       & Test    & 3.68            & --                &                    \\
                \cline{1-4}
                \multirow{2}*{TOPIC}   & Val     & 27.56           & --                &                    \\
                \cline{2-4}
                                       & Test    & 33.07           & --                &                    \\
                \cline{1-4}
                \multirow{2}*{LPT}     & Val     & 38.62           & --                &                    \\
                \cline{2-4}
                                       & Test    & 46.48           & --                &                    \\
                \hline
                \multirow{2}*{OCSORT}  & Val     & \textbf{439.84} & --                & \multirow{4}*{CPU} \\
                \cline{2-4}
                                       & Test    & \textbf{522.78} & --                &                    \\
                \cline{1-4}
                \multirow{2}*{HSG}     & Val     & 9.99            & \textbf{1.95}     &                    \\
                \cline{2-4}
                                       & Test    & 12.07           & \textbf{1.75}     &                    \\
                \Xhline{1.2pt}
            \end{tabular}
        \end{sloppypar}
    \end{center}
    \vspace{-.2cm}
\end{table}

Beyond qualitative analysis, we quantitatively evaluate the computational efficiency of our method against some association methods in Table~\ref{tab4_2}. Since comparisons are frame-based whereas our method is event-driven, we report two metrics, frames per second (FPS) and events per second (EPS). All evaluations are conducted on a platform equipped with an Intel i7-7800X CPU @3.50GHz and an NVIDIA RTX4090 GPU. The fastest method is OC-SORT. It achieves a significantly higher speed (over 10$\times$ faster) than the others. Although our method is slower than LPT, the other global optimization assignment method, it still processes over 1M EPS and requires no GPU resources. Given that the modeling stage with $O(\mathcal{N}^2)$ is the primary computational bottleneck, our method presents potential for further acceleration via parallelization.

\begin{table}
    \begin{center}
        \renewcommand\arraystretch{1.35}
        \begin{sloppypar}
            \vspace{-.2cm}
            \setlength{\abovecaptionskip}{-0.15cm}
            \caption{Ablation of B\'{e}zier curve-based term (\ding{192}), and direction-displacement-based term (\ding{193}) under YOLOX detections on eTram. ``w/o'' means without the corresponding component. }
            \vspace{0.1cm}
            \label{tab4}
            \begin{tabular}{
                >{\centering}p{1.6cm}|
                >{\centering}p{1.4cm}|
                >{\centering}p{0.7cm}|
                >{\centering}p{0.7cm}|
                >{\centering}p{0.9cm}|
                >{\centering\arraybackslash}p{0.9cm}
                }
                \Xhline{1.2pt}
                \multirow{2}*{Method}         & \multirow{2}*{Dataset} & MT            & ML            & MOTA          & IDF1          \\
                                              &                        & (\%)          & (\%)          & (\%)          & (\%)          \\
                \Xhline{1.2pt}
                \multirow{2}*{w/o \ding{192}} & Val                    & 42.2          & 23.2          & \textbf{27.5} & {52.5}        \\
                \cline{2-6}
                                              & Test                   & 16.7          & 39.4          & 29.0          & 46.7          \\
                \hline
                \multirow{2}*{w/o \ding{193}} & Val                    & 42.3          & 23.2          & 27.3          & \textbf{52.7} \\
                \cline{2-6}
                                              & Test                   & 16.7          & 39.4          & 29.0          & 46.8          \\
                \hline
                \multirow{2}*{HSG}            & Val                    & \textbf{42.3} & \textbf{23.2} & 27.3          & 52.6          \\
                \cline{2-6}
                                              & Test                   & \textbf{16.7} & \textbf{39.4} & \textbf{29.0} & \textbf{47.2} \\
                \Xhline{1.2pt}
            \end{tabular}
        \end{sloppypar}
    \end{center}
    \vspace{-.4cm}
\end{table}

\begin{table}
    \begin{center}
        \renewcommand\arraystretch{1.35}
        \begin{sloppypar}
            \setlength{\abovecaptionskip}{-0.15cm}
            \caption{Ablation of B\'{e}zier curve-based term (\ding{192}) and direction-displacement-based term (\ding{193}) on Ev-UAV. }
            \vspace{.1cm}
            \label{tab5}
            \begin{tabular}{
                >{\centering}p{1.3cm}|
                >{\centering}p{1.3cm}|
                >{\centering}p{1.3cm}|
                >{\centering}p{1.3cm}|
                >{\centering\arraybackslash}p{1.3cm}
                }
                \Xhline{1.2pt}
                \multirow{2}*{Method}         & \multirow{2}*{Dataset} & Track IoU     & Track Acc     & IDF1          \\
                                              &                        & (\%)          & (\%)          & (\%)          \\
                \Xhline{1.2pt}
                \multirow{2}*{w/o \ding{192}} & Val                    & 41.6          & 34.9          & 52.7          \\
                \cline{2-5}
                                              & Test                   & \textbf{51.6} & {43.0}        & \textbf{59.6} \\
                \hline
                \multirow{2}*{w/o \ding{193}} & Val                    & 51.8          & 61.6          & 62.5          \\
                \cline{2-5}
                                              & Test                   & 46.8          & 49.6          & 56.3          \\
                \hline
                \multirow{2}*{HSG}            & Val                    & 51.1          & 62.3          & 66.2          \\
                \cline{2-5}
                                              & Test                   & {47.8}        & \textbf{49.6} & {57.3}        \\
                \Xhline{1.2pt}
                \multirow{2}*{w/o \ding{192}} & Val$*$                 & 33.5          & 32.9          & 37.5          \\
                \cline{2-5}
                                              & Test$*$                & 33.6          & 32.1          & 37.2          \\
                \hline
                \multirow{2}*{w/o \ding{193}} & Val$*$                 & 49.1          & 63.3          & \textbf{83.0} \\
                \cline{2-5}
                                              & Test$*$                & 45.5          & 52.6          & \textbf{75.6} \\
                \hline
                \multirow{2}*{HSG}            & Val$*$                 & \textbf{49.9} & \textbf{63.7} & {81.3}        \\
                \cline{2-5}
                                              & Test$*$                & \textbf{47.2} & \textbf{52.7} & {74.0}        \\
                \Xhline{1.2pt}
            \end{tabular}
        \end{sloppypar}
    \end{center}
\end{table}

\subsection{Ablation Studies}
\label{section:ablation}
Our ablation studies focus on two modeling terms of our fuzzy modeling pipeline and the adaptive threshold $\tau$ in the graph partitioning. Experimental settings remain identical to before. Ablation results on modeling terms are presented in Table~\ref{tab4} and Table~\ref{tab5}. Besides, we conduct a quantitative analysis of the B\'{e}zier curve order, comparing linear, quadratic, and cubic fitting, as shown in Table~\ref{tab4_1} and Table~\ref{tab5_1}.

(1) \textit{Analysis of B\'{e}zier curve-based position smoothness}.
On eTram, the improvement brought by the B\'{e}zier curve-based position smoothness is modest, as object motion is predominantly linear, making the line-based error metric the primary distinguishing factor. In contrast, on Ev-UAV, where tracklets undergo large spatio-temporal variations, the B\'{e}zier curve model becomes the preferred choice. On the original set, the performance of our method is lower than that of the variant without \ding{192}. This occurs because the incorrect GT tracking labels divide an object trajectory into a few segments, further penalizing our metrics as one predicted trajectory can only match one GT trajectory. However, on the corrected set, our method yields nearly 2$\times$ improvements over the variant without this term, clearly demonstrating its effectiveness in handling complex motion patterns and long-range association.

\begin{table}
    \begin{center}
        \renewcommand\arraystretch{1.35}
        \begin{sloppypar}
            \vspace{-.2cm}
            \setlength{\abovecaptionskip}{-0.15cm}
            \caption{Ablation of linear (\ding{192}-l), quadratic (\ding{192}), and cubic (\ding{192}-c) B\'{e}zier curve-based term under YOLOX detections on eTram.}
            \vspace{0.1cm}
            \label{tab4_1}
            \begin{tabular}{
                >{\centering}p{1.6cm}|
                >{\centering}p{1.4cm}|
                >{\centering}p{0.7cm}|
                >{\centering}p{0.7cm}|
                >{\centering}p{0.9cm}|
                >{\centering\arraybackslash}p{0.9cm}
                }
                \Xhline{1.2pt}
                \multirow{2}*{Method}         & \multirow{2}*{Dataset} & MT            & ML            & MOTA          & IDF1          \\
                                              &                        & (\%)          & (\%)          & (\%)          & (\%)          \\
                \Xhline{1.2pt}

                \multirow{2}*{w \ding{192}-l} & Val                    & 42.3          & 23.2          & {27.2}        & {52.1}        \\
                \cline{2-6}
                                              & Test                   & 16.7          & 39.4          & 29.0          & 46.9          \\
                \hline
                \multirow{2}*{w \ding{192}-c} & Val                    & 42.3          & 23.2          & {27.2}        & {52.1}        \\
                \cline{2-6}
                                              & Test                   & 16.7          & 39.4          & 28.9          & 46.7          \\
                \hline
                \multirow{2}*{HSG}            & Val                    & \textbf{42.3} & \textbf{23.2} & \textbf{27.3} & \textbf{52.6} \\
                \cline{2-6}
                                              & Test                   & \textbf{16.7} & \textbf{39.4} & \textbf{29.0} & \textbf{47.2} \\
                \Xhline{1.2pt}
            \end{tabular}
        \end{sloppypar}
    \end{center}
    \vspace{-.2cm}
\end{table}

\begin{table}
    \begin{center}
        \renewcommand\arraystretch{1.35}
        \begin{sloppypar}
            \setlength{\abovecaptionskip}{-0.15cm}
            \caption{Ablation of the linear (\ding{192}-l), quadratic (\ding{192}), and cubic (\ding{192}-c) B\'{e}zier curve-based term on Ev-UAV. }
            \vspace{.15cm}
            \label{tab5_1}
            \begin{tabular}{
                >{\centering}p{1.3cm}|
                >{\centering}p{1.3cm}|
                >{\centering}p{1.3cm}|
                >{\centering}p{1.3cm}|
                >{\centering\arraybackslash}p{1.3cm}
                }
                \Xhline{1.2pt}
                \multirow{2}*{Method}         & \multirow{2}*{Dataset} & Track IoU     & Track Acc     & IDF1          \\
                                              &                        & (\%)          & (\%)          & (\%)          \\
                \Xhline{1.2pt}
                \multirow{2}*{w \ding{192}-l} & Val                    & \textbf{54.9} & 64.9          & \textbf{68.1} \\
                \cline{2-5}
                                              & Test                   & 42.4          & 47.6          & 59.7          \\
                \hline
                \multirow{2}*{w \ding{192}-c} & Val                    & 51.4          & \textbf{67.4} & 60.5          \\
                \cline{2-5}
                                              & Test                   & 40.1          & 48.3          & \textbf{59.9} \\
                \hline
                \multirow{2}*{HSG}            & Val                    & 51.1          & 62.3          & 66.2          \\
                \cline{2-5}
                                              & Test                   & \textbf{47.8} & \textbf{49.6} & {57.3}        \\
                \Xhline{1.2pt}
                \multirow{2}*{w \ding{192}-l} & Val  $*$               & \textbf{54.5} & \textbf{67.1} & {81.2}        \\
                \cline{2-5}
                                              & Test$*$                & 46.9          & 51.8          & 69.1          \\
                \hline
                \multirow{2}*{w \ding{192}-c} & Val $*$                & 48.7          & 62.2          & 77.4          \\
                \cline{2-5}
                                              & Test$*$                & 40.6          & 50.5          & \textbf{75.4} \\

                \hline
                \multirow{2}*{HSG}            & Val$*$                 & {49.9}        & {63.7}        & \textbf{81.3} \\
                \cline{2-5}
                                              & Test$*$                & \textbf{47.2} & \textbf{52.7} & {74.0}        \\
                \Xhline{1.2pt}
            \end{tabular}
        \end{sloppypar}
    \end{center}
\end{table}

(2) \textit{Analysis of the B\'{e}zier curve order}.
On eTram, our selected quadratic B\'{e}zier curve consistently outperforms the linear and cubic counterparts by 0.1\%-0.3\% in terms of MOTA and IDF1. On Ev-UAV, the quadratic curve model achieves the best trade-off. This experiment validates our analysis of B\'{e}zier curves in Sec.~\ref{section:PS}.

\begin{figure}[!t]
    \begin{center}
        \centerline{\includegraphics[width=.95\linewidth]{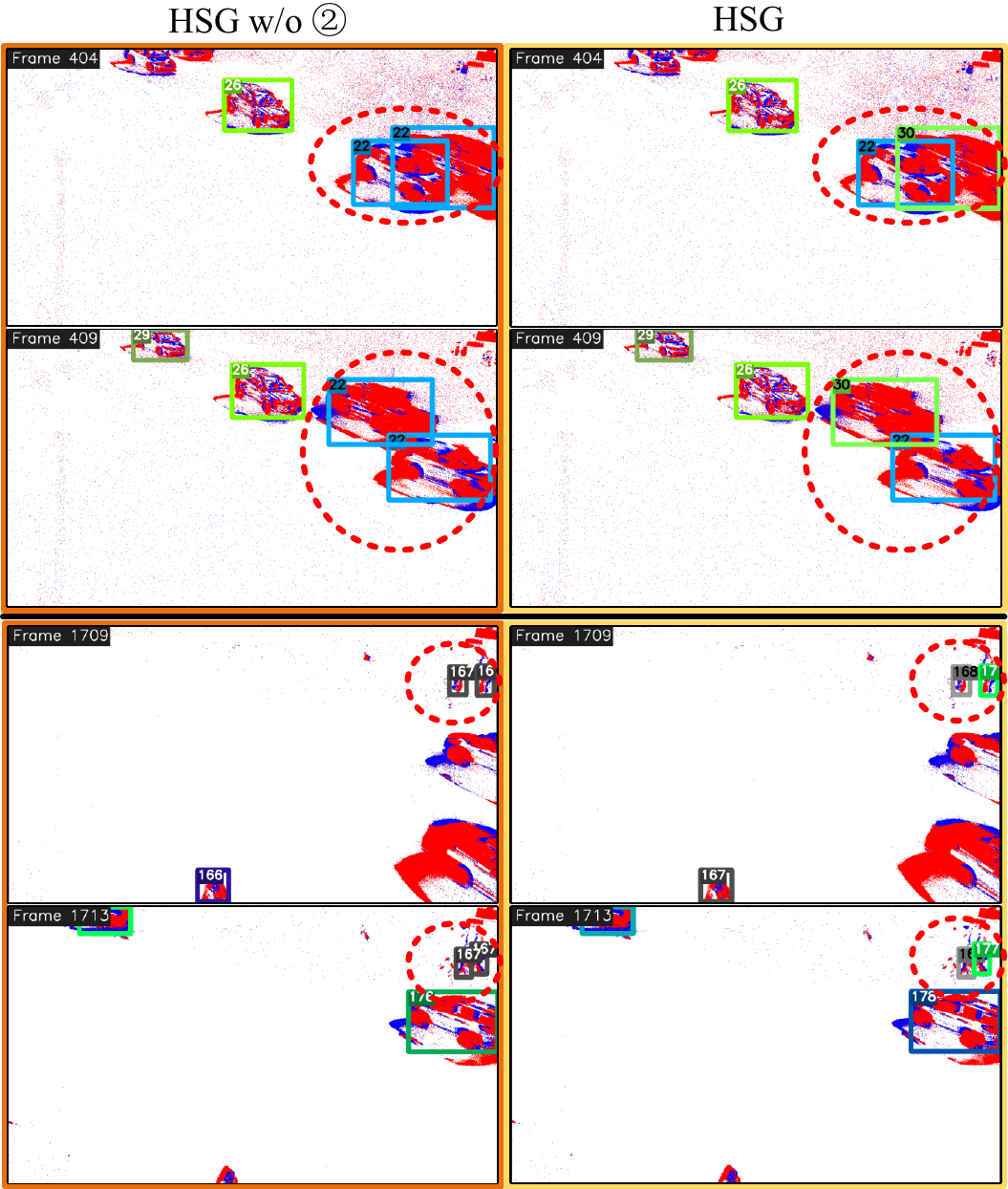}}
        \vspace{-.2cm}
        \caption{Ablation of the direction-displacement-based variation consistency term (\ding{193}) on eTram. Red dashed ellipses highlight cases where multiple tracklets have close positions but different moving directions. Two sequences are presented, separated by a black line: test\_day\_001 (top) and test\_day\_005 (bottom).}
        \label{fig5_5}
    \end{center}
    \vspace{-.5cm}
\end{figure}

(3) \textit{Analysis of direction-displacement-based variation consistency}. On eTram, the variant without this term achieves a slightly higher IDF1 than HSG by $0.1\%$. This marginal gain stems from a reduction in tracklet fragmentation, but as analyzed in Sec.~\ref{section:DS}, it also reflects over-association in some cases. To further analyze this, we visualize several test cases in Fig.~\ref{fig5_5}, where tracklets needed to be distinguished are highlighted by red dashed ellipses. HSG correctly associates them, whereas the variant without this term incorrectly associates them, confirming that this term is essential for distinguishing tracklets in such cases. The results on Ev-UAV also validate the importance of this term since the full model performs better than the variant without this term most of the time.

(4) \textit{Analysis of temporal weight $\lambda$}. As shown in Table~\ref{tab8}, omitting $\lambda$ drops IDF1 by over 15\% under GT detections and 10\% under YOLOX detections. This degradation confirms that $\lambda$ effectively mitigates the association unreliability introduced by long time intervals, thereby ensuring accurate tracking performance. Detailed factor ablations and sensitivity analyses are deferred to the supplementary material.

\begin{table}[!t]
    \begin{center}
        \renewcommand\arraystretch{1.35}
        \begin{sloppypar}
            \setlength{\abovecaptionskip}{-0.15cm}
            \caption{Ablations of $\lambda$ on eTram.}
            \label{tab8}
            \vspace{0.1cm}
            \begin{tabular}{
                >{\centering}p{1.2cm}|
                >{\centering}p{1.2cm}|
                >{\centering}p{1.cm}|
                >{\centering}p{1.5cm}|
                >{\centering\arraybackslash}p{1.6cm}
                }
                \Xhline{1.2pt}
                {Method}                     & {Detections}         & {Dataset} & MOTA    (\%)  & IDF1  (\%)    \\
                \Xhline{1.2pt}
                \multirow{2}*{w/o $\lambda$} & \multirow{4}*{GT}    & Val       & 98.2          & 80.1          \\
                \cline{3-5}
                                             &                      & Test      & 98.0          & 77.1          \\
                \cline{1-1}\cline{3-5}
                \multirow{2}*{HSG}           &                      & Val       & \textbf{98.5} & \textbf{95.7} \\
                \cline{3-5}
                                             &                      & Test      & \textbf{98.4} & \textbf{95.3} \\
                \Xhline{1.2pt}

                \multirow{2}*{w/o $\lambda$} & \multirow{4}*{YOLOX} & Val       & 98.2          & 80.1          \\
                \cline{3-5}
                                             &                      & Test      & 98.0          & 77.1          \\
                \cline{1-1}\cline{3-5}
                \multirow{2}*{HSG}           &                      & Val       & \textbf{98.5} & \textbf{95.7} \\
                \cline{3-5}
                                             &                      & Test      & \textbf{98.4} & \textbf{95.3} \\
                \Xhline{1.2pt}
            \end{tabular}
        \end{sloppypar}
    \end{center}
\end{table}

\begin{table}
    \begin{center}
        \renewcommand\arraystretch{1.35}
        \begin{sloppypar}
            \setlength{\abovecaptionskip}{-0.15cm}
            \caption{Ablation of factors in deriving $\tau$, including node size balancing factor $\gamma$ (\ding{194}) and denominator $m+1$ (\ding{195}) under YOLOX detections on eTram. $*$ and $^{\ddagger}$ denote statistically significant improvement ($p < 0.05$) over \ding{194} and \ding{195} (paired t-test, $n{=}25$ for Val and $n{=}30$ for Test), respectively.}
            \vspace{.1cm}
            \label{tab6}
            \begin{tabular}{
                >{\centering}p{.7cm}|
                >{\centering}p{.7cm}|
                >{\centering}p{0.9cm}|
                >{\centering}p{0.9cm}|
                >{\centering}p{1.5cm}|
                >{\centering\arraybackslash}p{1.5cm}
                }
                \Xhline{1.2pt}
                {Method}                      & {Dataset} & MT(\%)/p      & ML(\%)/p      & MOTA(\%)/p                  & IDF1(\%)/p                  \\
                \Xhline{1.2pt}
                \multirow{2}*{w/o \ding{194}} & Val       & 42.3/--       & 23.2/--       & 27.2/1.21E-2                & {52.0/3.99E-2}              \\
                \cline{2-6}
                                              & Test      & 16.7/--       & 39.4/--       & 29.0/3.56E-4                & 46.9/2.18E-2                \\
                \hline
                \multirow{2}*{w/o \ding{195}} & Val       & 42.3/--       & 23.2/--       & 27.0/4.01E-9                & 47.2/1.77E-5                \\
                \cline{2-6}
                                              & Test      & 16.7/--       & 39.4/--       & 28.8/7.24E-8                & 42.0/6.25E-6                \\
                \hline
                \multirow{2}*{HSG}            & Val       & \textbf{42.3} & \textbf{23.2} & \textbf{27.3}$^{*\ddagger}$ & \textbf{52.6}$^{*\ddagger}$ \\
                \cline{2-6}
                                              & Test      & \textbf{16.7} & \textbf{39.4} & \textbf{29.0}$^{*\ddagger}$ & \textbf{47.2}$^{*\ddagger}$ \\
                \Xhline{1.2pt}
            \end{tabular}
        \end{sloppypar}
    \end{center}
\end{table}

\begin{table}
    \begin{center}
        \renewcommand\arraystretch{1.35}
        \begin{sloppypar}
            \setlength{\abovecaptionskip}{-0.15cm}
            \caption{Ablation of factors in deriving $\tau$, including the balancing factor $\gamma$ (\ding{194}) and the denominator $m+1$ (\ding{195}) on Ev-UAV.}
            \vspace{.1cm}
            \label{tab7}
            \begin{tabular}{
                >{\centering}p{1.3cm}| 
                >{\centering}p{1.3cm}|
                >{\centering}p{1.3cm}|
                >{\centering}p{1.3cm}|
                >{\centering\arraybackslash}p{1.3cm}
                }
                \Xhline{1.2pt}
                \multirow{2}*{Method}         & \multirow{2}*{Dataset} & Track IoU     & Track Acc     & IDF1          \\
                                              &                        & (\%)          & (\%)          & (\%)          \\
                \Xhline{1.2pt}
                \multirow{2}*{w/o \ding{194}} & Val                    & 51.4          & 62.5          & {67.2}        \\
                \cline{2-5}
                                              & Test                   & 47.8          & 49.6          & 57.3          \\
                \hline
                \multirow{2}*{w/o \ding{195}} & Val                    & \textbf{53.1} & \textbf{64.2} & \textbf{68.7} \\
                \cline{2-5}
                                              & Test                   & {47.1}        & \textbf{50.3} & \textbf{58.8} \\
                \hline
                \multirow{2}*{HSG}            & Val                    & 51.1          & 62.3          & 66.2          \\
                \cline{2-5}
                                              & Test                   & \textbf{47.8} & {49.6}        & {57.3}        \\
                \Xhline{1.2pt}
                \multirow{2}*{w/o \ding{194}} & Val$*$                 & \textbf{50.3} & \textbf{64.0} & {82.5}        \\
                \cline{2-5}
                                              & Test$*$                & 47.2          & 52.7          & 74.0          \\
                \hline
                \multirow{2}*{w/o \ding{195}} & Val$*$                 & 47.5          & 63.4          & \textbf{86.6} \\
                \cline{2-5}
                                              & Test$*$                & \textbf{48.3} & \textbf{53.6} & 72.9          \\
                \hline
                \multirow{2}*{HSG}            & Val$*$                 & {49.9}        & {63.7}        & {81.3}        \\
                \cline{2-5}
                                              & Test$*$                & {47.2}        & {52.7}        & \textbf{74.0} \\
                \Xhline{1.2pt}
            \end{tabular}
        \end{sloppypar}
    \end{center}
\end{table}

\begin{figure}[!t]
    \begin{center}
        \centerline{\includegraphics[width=1\linewidth]{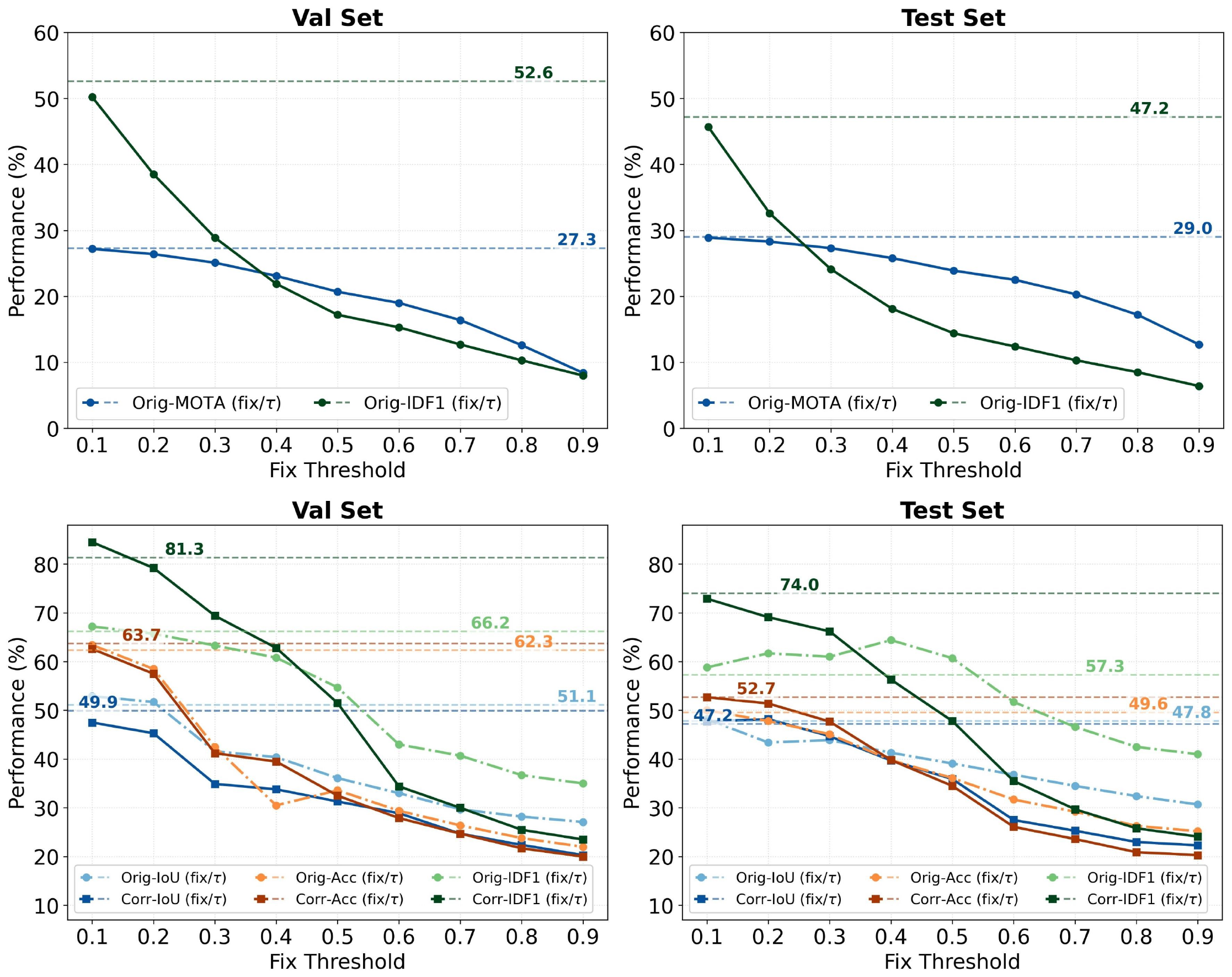}}
        \vspace{-0.2cm}
        \caption{Comparisons of adaptive merging threshold $\tau$ with a series of fixed thresholds. The first row is the results on eTram, and the second row is the results on Ev-UAV.}
        \label{fig5_6}
    \end{center}
    \vspace{-.3cm}
\end{figure}

(5) \textit{Analysis of $\tau$}. We evaluate the effectiveness of $\tau$ and the necessity of its factor $\gamma$ and denominator $m+1$. Fig.~\ref{fig5_6} compares our adaptive threshold $\tau$ against a series of fixed thresholds ranging from 0.1 to 0.9. The performance using $\tau$ is represented by horizontal dashed lines. Evidently, using $\tau$ achieves more robust and accurate tracking than the fixed thresholds. Table~\ref{tab6} and Table~\ref{tab7} report the performance of our method without $\gamma$ or the denominator $m+1$ from the $\tau$ calculation. On eTram, our full method consistently outperforms these variants. To validate the improvements, we also conducted paired t-tests on per-sequence metrics. As shown in Table~\ref{tab6}, HSG yields statistically significant improvements over the two variants with $p<0.05$. In contrast to this, two variants perform on par with or even better than our full method on Ev-UAV. We attribute this to the differing graph sparsity across datasets. Specifically, $\gamma$ and $m+1$ correction are designed for dense graphs, addressing size imbalance and overly strict merging conditions, respectively. When the HSG is sparse, the connectivity within a vertex degenerates into a path, and their effect diminishes or even introduces few incorrect associations. These results confirm that adapting the method's configuration to the target scenario can improve performance across different applications.

\subsection{Discussion}
Experiments demonstrate the effectiveness and superiority of our method. However, our approach still possesses certain limitations. Fundamentally, our graph partitioning is akin to a voting system, assigning each tracklet to the most likely trajectory without enforcing global continuity within a partition. This can lead to erroneous associations when voting evidence is insufficient or ambiguous. For example, under sparse evidence, an isolated tracklet may be incorrectly absorbed into a continuous trajectory merely due to spatial proximity and a similar motion direction. Conversely, when evidence is ambiguous, such as in extremely dense crowds, the limitation becomes equally pronounced. In such scenarios, event trajectories of adjacent targets heavily overlap and exhibit similar motion patterns, introducing severe ambiguity into the voting mechanism. Addressing these limitations opens up several future directions, including examining a sliding window of tracklets within each partition after each round of graph partitioning, or designing incremental metrics that capture motion continuity across multiple tracklets.

\section{Conclusion}
\label{section:Con}
In this work, we formulate the event-based multi-dimensional assignment problem as a graph partition task. Unlike image-based tracking, event-based tracking inherently involves the many-to-many association, which is an underdetermined and NP-hard task arising from the asynchronous nature of events. Therefore, we propose a fuzzy modeling method to construct a Hierarchical Sparse Graph (HSG) that captures relationships between candidate pairs. Subsequently, an automatic graph partitioning algorithm is also developed based on the HSG to obtain optimal assignments. The overall method is training-free and plug-and-play.
Experiments on both one-to-one and many-to-many matching scenarios show the strong performance of our method across diverse challenging conditions, particularly in long-range and tiny object tracking.

\vspace{-1cm}
\begin{IEEEbiography}[{\includegraphics[width=1in,height=1.25in,clip,keepaspectratio]{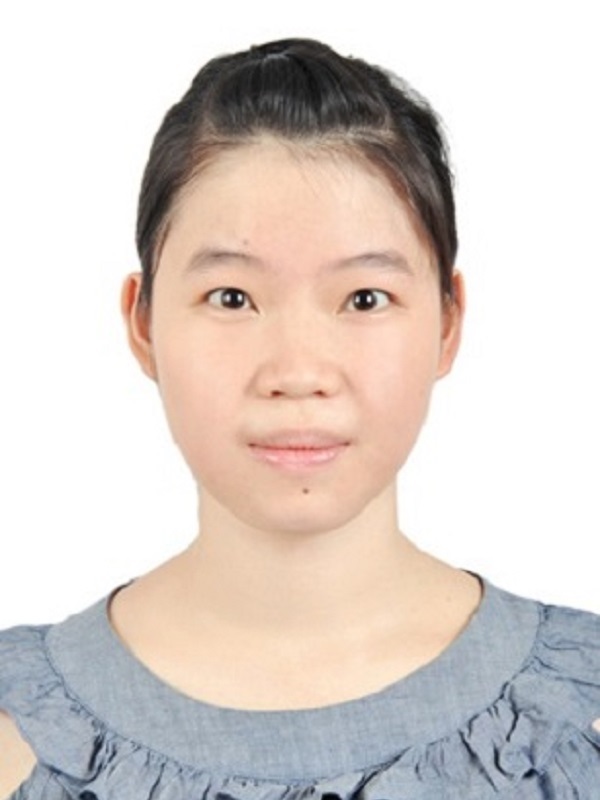}}]{Muxi Zha}
    received the B.E. degree from University of Science \& Technology Beijing, Beijing, China in 2020, and the M.E. degree from the College of Electrical and Information Engineering, Hunan University, Changsha, China, in 2023. She is currently pursuing the Ph.D. degree with the College of Aerospace Science and Engineering, National University of Defense Technology, Changsha, China. Her research interests include data processing and object tracking in both image and event domains.
\end{IEEEbiography}
\vspace{-1.2cm}

\begin{IEEEbiography}[{\includegraphics[width=1in,height=1.25in,clip,keepaspectratio]{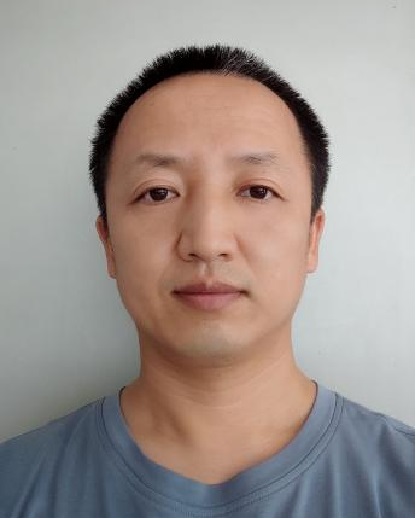}}]{Yang Shang}
    received the B.S. degree in aerodynamics, and the Ph.D. degree in aeronautical and astronautical science and technology from the National University of Defense Technology, Changsha, China, in 2000 and 2006, respectively. He is currently a Professor with the College of Aerospace Science and Engineering, National University of Defense Technology. His main research interests include photogrammetry, machine vision, and navigation systems.
\end{IEEEbiography}
\vspace{-1.2cm}

\begin{IEEEbiography}[{\includegraphics[width=1in,height=1.25in,clip,keepaspectratio]{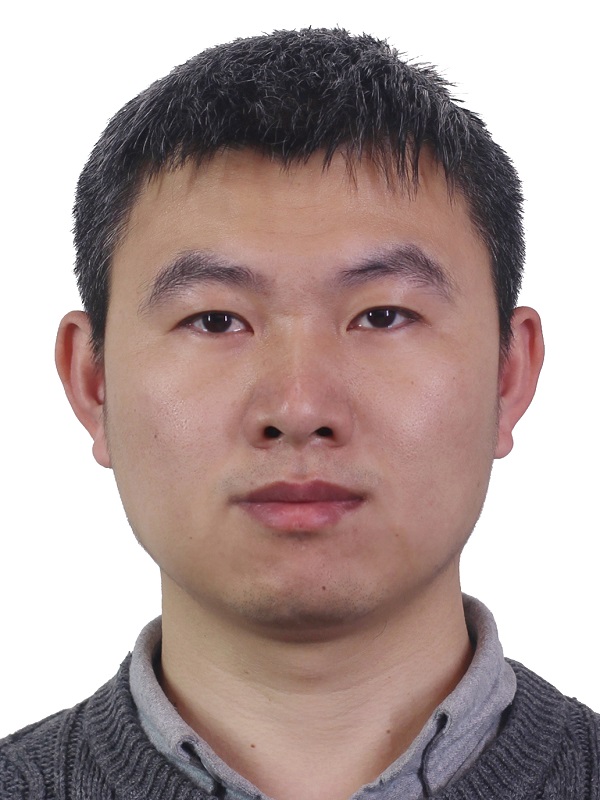}}]{Wenlong Zhang}
    received the B.E., M.S., and Ph.D. degrees from the National University of Defense Technology, Changsha, China, in 2014, 2016, and 2020, respectively. He is currently a Lecturer with the College of Aerospace Science and Engineering, National University of Defense Technology. His research interests include image processing and compute vision.\end{IEEEbiography}
\vspace{-1.1cm}

\begin{IEEEbiography}[{\includegraphics[width=1in,height=1.25in,clip,keepaspectratio]{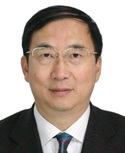}}]{Qifeng Yu}
    received the B.S. degree from Northwestern Polytechnic University, Xian, China, in 1981, the M.S. degree from the National University of
    Defense Technology, Changsha, China, in 1984, and the Ph.D. degree from Bremen University, Bremen, Germany, in 1996. Currently, he is a Professor at the National University of Defense Technology. He is a Member of the Chinese Academy of Sciences. His current research fields are image measurement and vision navigation.
\end{IEEEbiography}
\vspace{-1.2cm}

\begin{IEEEbiography}[{\includegraphics[width=1in,height=1.25in,clip,keepaspectratio]{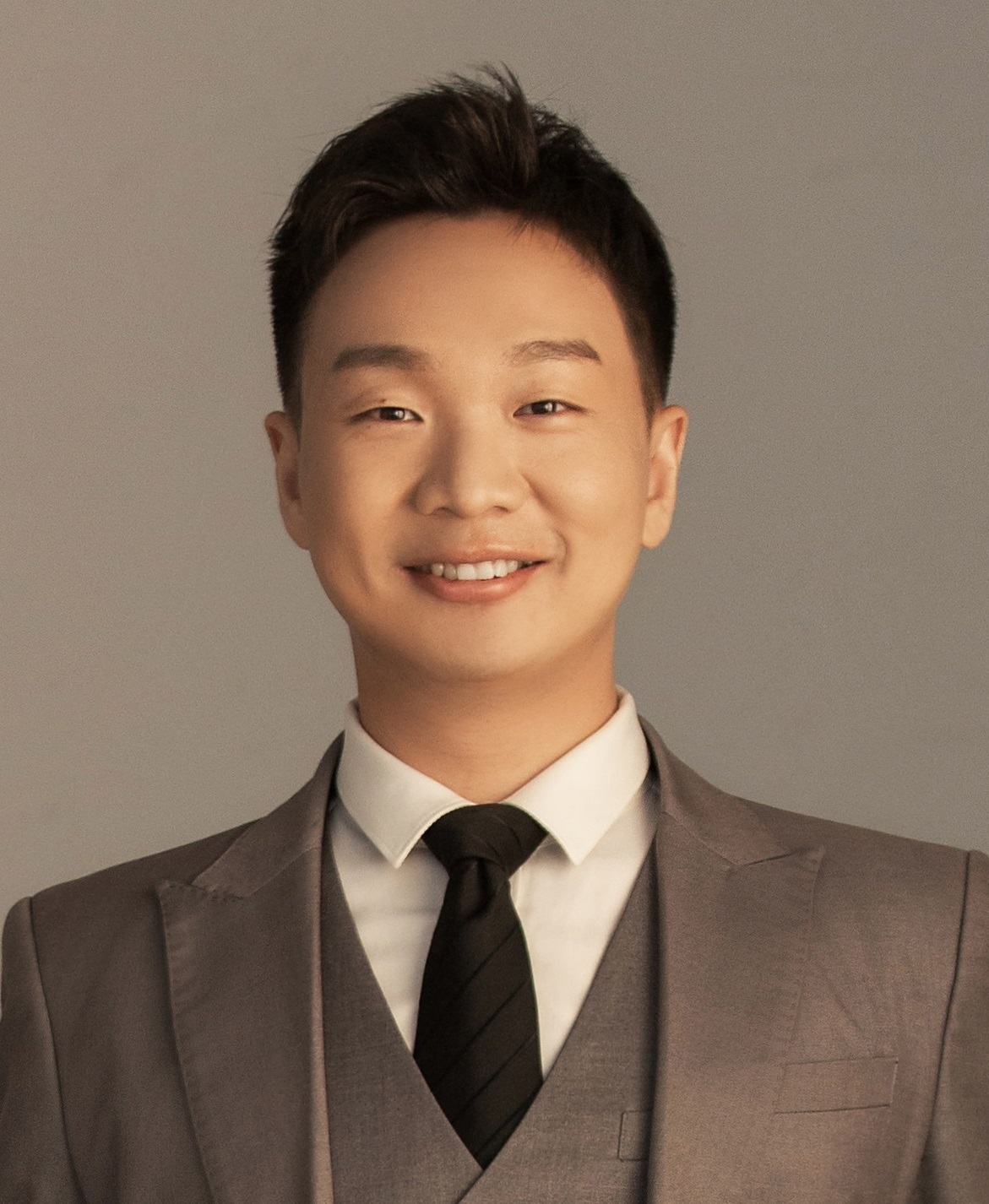}}]{Banglei Guan}
    received a B.S. degree in Geomatics Engineering from Wuhan University, Wuhan, China, in 2012 and a PhD degree in Aeronautical and Astronautical Science and Technology from National University of Defense Technology, Changsha, China, in 2018. From 2016 to 2017, he was an Exchange Student at Graz University of Technology, Graz, Austria. He is currently a professor with the National University of Defense Technology. His research interests include photomechanics and videometrics.
\end{IEEEbiography}

\balance

\vfill

\begin{thebibliography}{99}

    \bibitem{baeRobustOnlineMultiobject2014}
    S.~H. Bae and K.~J. Yoon, ``Robust {{Online Multiobject Tracking With Data Association}} and {{Track Management}},'' \emph{IEEE Trans. Image Process.}, vol.~23, no.~7, pp. 2820--2833, Jul. 2014.

    \bibitem{gaoRobustTrackingLearning2023}
    J.~Gao, B.~Zhong, and Y.~Chen, ``Robust {{Tracking}} via {{Learning Model Update With Unsupervised Anomaly Detection Philosophy}},'' \emph{IEEE Trans. Circuits Syst. Video Technol.}, vol.~33, no.~5, pp. 2330--2341, May 2023.

    \bibitem{rakaiDataAssociationMultiple2022}
    L.~Rakai, H.~Song, S.~Sun, W.~Zhang, and Y.~Yang, ``Data association in multiple object tracking: {{A}} survey of recent techniques,'' \emph{Expert Syst. Appl.}, vol. 192, p. 116300, Apr. 2022.

    \bibitem{wangMambaEVTEventStreamBased2026}
    X.~Wang, C.~Wang, S.~Wang, X.~Wang, Z.~Zhao, L.~Zhu, and B.~Jiang, ``{{MambaEVT}}: {{Event Stream-Based Visual Object Tracking Using State Space Model}},'' \emph{IEEE Transactions on Circuits and Systems for Video Technology}, vol.~36, no.~1, pp. 278--291, Jan. 2026.

    \bibitem{tokmakovLearningTrackObject2021}
    P.~Tokmakov, J.~Li, W.~Burgard, and A.~Gaidon, ``Learning to {{Track}} with {{Object Permanence}},'' in \emph{Proc. IEEE/CVF Int. Conf. Comput. Vis.}, Montreal, QC, Canada, Oct. 2021, pp. 10\,840--10\,849.

    \bibitem{liuMultiObjectTrackingMeets2022}
    S.~Liu, X.~Li, H.~Lu, and Y.~He, ``Multi-{{Object Tracking Meets Moving UAV}},'' in \emph{Proc. IEEE/CVF Conf. Comput. Vis. Pattern Recognit.}, New Orleans, Louisiana, USA, 2022.

    \bibitem{caoObservationCentricSORTRethinking2023}
    J.~Cao, J.~Pang, X.~Weng, R.~Khirodkar, and K.~Kitani, ``Observation-{{Centric SORT}}: {{Rethinking SORT}} for {{Robust Multi-Object Tracking}},'' in \emph{Proc. IEEE/CVF Conf. Comput. Vis. Pattern Recognit.}, Vancouver, Canada, 2023, pp. 9686--9696.

    \bibitem{zhangGlobalDataAssociation2008}
    L.~Zhang, Y.~Li, and R.~Nevatia, ``Global data association for multi-object tracking using network flows,'' in \emph{Proc. IEEE/CVF Conf. Comput. Vis. Pattern Recognit.}, Anchorage, Alaska, Jun. 2008, pp. 1--8.

    \bibitem{linLTTrackRethinkingTracking2024}
    J.~Lin, G.~Liang, and R.~Zhang, ``{{LTTrack}}: {{Rethinking}} the {{Tracking Framework}} for {{Long-Term Multi-Object Tracking}},'' \emph{IEEE Trans. Circuits Syst. Video Technol.}, vol.~34, no.~10, pp. 9866--9881, Oct. 2024.

    \bibitem{buttMultitargetTrackingLagrangian2013}
    A.~A. Butt and R.~T. Collins, ``Multi-target {{Tracking}} by {{Lagrangian Relaxation}} to {{Min-cost Network Flow}},'' in \emph{Proc. IEEE/CVF Conf. Comput. Vis. Pattern Recognit.}, Jun. 2013, pp. 1846--1853.

    \bibitem{ochsHigherOrderMotion2012}
    P.~Ochs and T.~Brox, ``Higher order motion models and spectral clustering,'' in \emph{Proc. IEEE/CVF Conf. Comput. Vis. Pattern Recognit.}, Providence, RI, USA, Jun. 2012, pp. 614--621.

    \bibitem{gehrigRecurrentVisionTransformers2023}
    M.~Gehrig and D.~Scaramuzza, ``Recurrent {{Vision Transformers}} for {{Object Detection With Event Cameras}},'' in \emph{Proc. IEEE/CVF Conf. Comput. Vis. Pattern Recognit.}, Vancouver, Canada, 2023, pp. 13\,884--13\,893.

    \bibitem{vermaETraMEventbasedTraffic2024}
    A.~A. Verma, B.~Chakravarthi, A.~Vaghela, H.~Wei, and Y.~Yang, ``{{eTraM}}: {{Event-based Traffic Monitoring Dataset}},'' in \emph{Proc. IEEE/CVF Conf. Comput. Vis. Pattern Recognit.}, Seattle WA, USA, 2024, pp. 22\,637--22\,646.

    \bibitem{chenEventbasedTinyObject2025}
    N.~Chen, C.~Xiao, Y.~Dai, S.~He, M.~Li, and W.~An, ``Event-based {{Tiny Object Detection}}: {{A Benchmark Dataset}} and {{Baseline}},'' in \emph{Proc. IEEE/CVF Int. Conf. Comput. Vis.}, Honolulu, Hawaii, Jun. 2025.

    \bibitem{gallegoEventbasedVisionSurvey2022}
    G.~Gallego, T.~Delbruck, G.~Orchard, C.~Bartolozzi, B.~Taba, A.~Censi, S.~Leutenegger, A.~Davison, J.~Conradt, K.~Daniilidis, and D.~Scaramuzza, ``Event-based {{Vision}}: {{A Survey}},'' \emph{IEEE Trans. Pattern Anal. Mach. Intell.}, vol.~44, no.~1, pp. 154--180, Jan. 2022.

    \bibitem{huangEventGuidedStructuredOutput2018a}
    J.~Huang, S.~Wang, M.~Guo, and S.~Chen, ``Event-{{Guided Structured Output Tracking}} of {{Fast-Moving Objects Using}} a {{CeleX Sensor}},'' \emph{IEEE Trans. Circuits Syst. Video Technol.}, vol.~28, no.~9, pp. 2413--2417, Sep. 2018.

    \bibitem{wangPriorKnowledgeDrivenHybrid2025}
    M.~Wang, F.~Shi, X.~Cheng, and S.~Chen, ``Prior {{Knowledge-Driven Hybrid Prompter Learning}} for {{RGB-Event Tracking}},'' \emph{IEEE Trans. Circuits Syst. Video Technol.}, vol.~35, no.~9, pp. 8679--8691, Sep. 2025.

    \bibitem{liuEMTrackEfficientMultimodal2025}
    C.~Liu, Z.~Guan, S.~Lai, Y.~Liu, H.~Lu, and D.~Wang, ``{{EMTrack}}: {{Efficient Multimodal Object Tracking}},'' \emph{IEEE Trans. Circuits Syst. Video Technol.}, vol.~35, no.~3, pp. 2202--2214, Mar. 2025.

    \bibitem{mengBRMoEBlindMultiModal2025}
    Q.~Meng, A.~Lu, and Z.~Jin, ``{{BR-MoE}}: {{Blind Multi-Modal Tracking With Route-Dynamic Mixture}} of {{Experts}},'' \emph{IEEE Trans. Circuits Syst. Video Technol.}, vol.~35, no.~12, pp. 11\,908--11\,921, Dec. 2025.

    \bibitem{dietschePowerlineTrackingEvent2021}
    A.~Dietsche, G.~Cioffi, J.~{Hidalgo-Carrio}, and D.~Scaramuzza, ``Powerline {{Tracking}} with {{Event Cameras}},'' in \emph{Proc. IEEE/RSJ Int. Conf. Intell. Rob. Syst.}\hskip 1em plus 0.5em minus 0.4em\relax Prague, Czech Republic: IEEE, Sep. 2021, pp. 6990--6997.

    \bibitem{everdingLowLatencyLineTracking2018}
    L.~Everding and J.~Conradt, ``Low-{{Latency Line Tracking Using Event-Based Dynamic Vision Sensors}},'' \emph{Front. Neurorobot.}, vol.~12, 2018.

    \bibitem{legentilIDOLFrameworkIMUDVS2020}
    C.~Le~Gentil, F.~Tschopp, I.~Alzugaray, T.~{Vidal-Calleja}, R.~Siegwart, and J.~Nieto, ``{{IDOL}}: {{A Framework}} for {{IMU-DVS Odometry}} using {{Lines}},'' in \emph{Proc. IEEE/RSJ Int. Conf. Intell. Rob. Syst.}\hskip 1em plus 0.5em minus 0.4em\relax Las Vegas, NV, USA: IEEE, Oct. 2020, pp. 5863--5870.

    \bibitem{zhuEventBasedVisualInertial2017}
    A.~Z. Zhu, N.~Atanasov, and K.~Daniilidis, ``Event-{{Based Visual Inertial Odometry}},'' in \emph{Proc. IEEE/CVF Conf. Comput. Vis. Pattern Recognit.}\hskip 1em plus 0.5em minus 0.4em\relax Honolulu, Hawaii: IEEE, Jul. 2017, pp. 5816--5824.

    \bibitem{jiangObjectTrackingEvent2020}
    R.~Jiang, X.~Mou, S.~Shi, Y.~Zhou, Q.~Wang, M.~Dong, and S.~Chen, ``Object tracking on event cameras with offline-online learning,'' \emph{CAAI Trans. Intell. Technol.}, vol.~5, no.~3, pp. 165--171, 2020.

    \bibitem{wangEventBlobTracking2023}
    Z.~Wang, T.~Molloy, P.~{van Goor}, and R.~Mahony, ``Event {{Blob Tracking}}: {{An Asynchronous Real-Time Algorithm}},'' \emph{arXiv preprint arXiv:2307.10593}, 2023.

    \bibitem{barranco2018real}
    F.~Barranco, C.~Fermuller, and E.~Ros, ``Real-time clustering and multi-target tracking using event-based sensors,'' in \emph{IEEE/RSJ Int. Conf. Intell. Robots Syst.}, 2018, pp. 5764--5769.

    \bibitem{alzugarayAsynchronousCornerDetection2018}
    I.~Alzugaray and M.~Chli, ``Asynchronous {{Corner Detection}} and {{Tracking}} for {{Event Cameras}} in {{Real Time}},'' \emph{IEEE Robot. Autom. Lett.}, vol.~3, no.~4, pp. 3177--3184, Oct. 2018.

    \bibitem{lagorceAsynchronousEventBasedMultikernel2015}
    X.~Lagorce, C.~Meyer, S.-H. Ieng, D.~Filliat, and R.~Benosman, ``Asynchronous {{Event-Based Multikernel Algorithm}} for {{High-Speed Visual Features Tracking}},'' \emph{IEEE Trans. Neural Netw. Learn. Syst.}, vol.~26, no.~8, 2015.

    \bibitem{alzugarayHASTEMultiHypothesisAsynchronous2020}
    I.~Alzugaray, ``{{HASTE}}: Multi-{{Hypothesis Asynchronous Speeded-up Tracking}} of {{Events}},'' in \emph{Proc. Brit. Mach. Vis. Conf.}, UK, 2020.

    \bibitem{hu2022ecdt}
    S.~Hu, Y.~Kim, H.~Lim, A.~J. Lee, and H.~Myung, ``ecdt: Event clustering for simultaneous feature detection and tracking,'' in \emph{IEEE/RSJ Int. Conf. Intell. Robots Syst.}, 2022, pp. 3808--3815.

    \bibitem{rameshETLDEventbasedFramework2020}
    B.~Ramesh, S.~Zhang, H.~Yang, A.~Ussa, M.~Ong, G.~Orchard, and C.~Xiang, ``E-{{TLD}}: {{Event-based Framework}} for {{Dynamic Object Tracking}},'' \emph{IEEE Trans. Circuits Syst. Video Technol.}, vol.~30, no.~10, pp. 3996--4006, Sep. 2020.

    \bibitem{chenAsynchronousTrackingDetectionAdaptive2019}
    H.~Chen, Q.~Wu, Y.~Liang, X.~Gao, and H.~Wang, ``Asynchronous {{Tracking-by-Detection}} on {{Adaptive Time Surfaces}} for {{Event-based Object Tracking}},'' in \emph{Proc. ACM Int. Conf. Multimedia}, France, 2019, pp. 473--481.

    \bibitem{chenEndtoEndLearningObject2020}
    H.~Chen, D.~Suter, Q.~Wu, and H.~Wang, ``End-to-{{End Learning}} of {{Object Motion Estimation}} from {{Retinal Events}} for {{Event-Based Object Tracking}},'' vol.~34, no.~07, Apr. 2020, pp. 10\,534--10\,541.

    \bibitem{berclazMultipleObjectTracking2011}
    J.~Berclaz, F.~Fleuret, E.~Turetken, and P.~Fua, ``Multiple {{Object Tracking Using K-Shortest Paths Optimization}},'' \emph{IEEE Trans. Pattern Anal. Mach. Intell.}, vol.~33, no.~9, pp. 1806--1819, Sep. 2011.

    \bibitem{pirsiavashGloballyoptimalGreedyAlgorithms2011}
    H.~Pirsiavash, D.~Ramanan, and C.~C. Fowlkes, ``Globally-optimal greedy algorithms for tracking a variable number of objects,'' in \emph{Proc. IEEE/CVF Conf. Comput. Vis. Pattern Recognit.}, Colorado Springs, CO, USA, Jun. 2011, pp. 1201--1208.

    \bibitem{collinsMultitargetDataAssociation2012}
    R.~T. Collins, ``Multitarget data association with higher-order motion models,'' in \emph{Proc. IEEE/CVF Conf. Comput. Vis. Pattern Recognit.}, Providence, RI, USA, Jun. 2012, pp. 1744--1751.

    \bibitem{poschmannFactorGraphBased2020}
    J.~P{\"o}schmann, T.~Pfeifer, and P.~Protzel, ``Factor {{Graph}} based {{3D Multi-Object Tracking}} in {{Point Clouds}},'' in \emph{Proc. IEEE/RSJ Int. Conf. Intell. Rob. Syst.}, Las Vegas, NV, USA, Oct. 2020, pp. 10\,343--10\,350.

    \bibitem{roshanzamirGMCPTrackerGlobalMultiobject2012a}
    A.~Roshan~Zamir, A.~Dehghan, and M.~Shah, ``{{GMCP-Tracker}}: {{Global Multi-object Tracking Using Generalized Minimum Clique Graphs}},'' in \emph{Proc. Eur. Conf. Comput. Vis.}\hskip 1em plus 0.5em minus 0.4em\relax Firenze, Italy: Springer, 2012, pp. 343--356.

    \bibitem{dehghanGMMCPTrackerGlobally2015}
    A.~Dehghan, S.~Modiri~Assari, and M.~Shah, ``{{GMMCP Tracker}}: {{Globally Optimal Generalized Maximum Multi Clique Problem}} for {{Multiple Object Tracking}},'' in \emph{Proc. IEEE/CVF Conf. Comput. Vis. Pattern Recognit.}, Boston, Massachusetts, USA, 2015, pp. 4091--4099.

    \bibitem{zhu2017event}
    A.~Z. Zhu, N.~Atanasov, and K.~Daniilidis, ``Event-based feature tracking with probabilistic data association,'' in \emph{Proc. IEEE Int. Conf. Robot. Autom.}, Shanghai, China, 2017, pp. 4465--4470.

    \bibitem{seokRobustFeatureTracking2020a}
    H.~Seok and J.~Lim, ``Robust {{Feature Tracking}} in {{DVS Event Stream}} using {{B{\'e}zier Mapping}},'' in \emph{Proc. IEEE/CVF Winter Conf. Appl. Comput. Vis.}, CO, U.S.A., Mar. 2020, pp. 1647--1656.

    \bibitem{chenAccurateEfficientApproximation2017}
    Y.~Chen, Y.~Cai, J.~Zheng, and D.~Thalmann, ``Accurate and {{Efficient Approximation}} of {{Clothoids Using B{\'e}zier Curves}} for {{Path Planning}},'' \emph{IEEE Trans. Robot.}, vol.~33, no.~5, pp. 1242--1247, Oct. 2017.

    \bibitem{lagorceHOTSHierarchyEventBased2016}
    X.~Lagorce, G.~Orchard, F.~Galluppi, B.~E. Shi, and R.~B. Benosman, ``{{HOTS}}: {{A Hierarchy}} of {{Event-Based Time-Surfaces}} for {{Pattern Recognition}},'' \emph{IEEE Trans. Pattern Anal. Mach. Intell.}, vol.~39, no.~7, pp. 1346--1359, Jul. 2016.

    \bibitem{blondelFastUnfoldingCommunities2008}
    V.~D. Blondel, J.~L. Guillaume, R.~Lambiotte, and E.~Lefebvre, ``Fast unfolding of communities in large networks,'' \emph{J. Stat. Mech.}, vol. 2008, no.~10, p. P10008, Oct. 2008.

    \bibitem{xuSCANStructuralClustering2007}
    X.~Xu, N.~Yuruk, Z.~Feng, and T.~A.~J. Schweiger, ``{{SCAN}}: A structural clustering algorithm for networks,'' in \emph{Proc. 13th ACM SIGKDD Int. Conf. Knowl. Discov. Data Min.}\hskip 1em plus 0.5em minus 0.4em\relax NY, USA: Association for Computing Machinery, Aug. 2007, pp. 824--833.

    \bibitem{newmanModularityCommunityStructure2006}
    M.~E.~J. Newman, ``Modularity and community structure in networks,'' \emph{Proc. Natl. Acad. Sci. USA}, vol. 103, no.~23, pp. 8577--8582, Jun. 2006.

    \bibitem{keniEvaluatingMultipleObject2008}
    B.~Keni and S.~Rainer, ``Evaluating {{Multiple Object Tracking Performance}}: {{The CLEAR MOT Metrics}},'' \emph{EURASIP J. Image Video Process.}, vol. 2008, no.~1, p. 246309, Jan. 2008.

    \bibitem{ristaniPerformanceMeasuresData2016b}
    E.~Ristani, F.~Solera, R.~Zou, R.~Cucchiara, and C.~Tomasi, ``Performance {{Measures}} and a {{Data Set}} for~{{Multi-target}}, {{Multi-camera Tracking}},'' in \emph{in Proc. Eur. Conf. Comput. Vis. Workshops}, G.~Hua and H.~J{\'e}gou, Eds.\hskip 1em plus 0.5em minus 0.4em\relax Cham: Springer International Publishing, 2016, pp. 17--35.

    \bibitem{mentastiEventBasedObjectDetection2022}
    S.~Mentasti, A.~W. Kambale, and M.~Matteucci, ``Event-{{Based Object Detection}} and {{Tracking}} - {{A Traffic Monitoring Use Case}} -,'' in \emph{Int. Conf. Image Anal. Process. Workshops}.\hskip 1em plus 0.5em minus 0.4em\relax Cham: Springer International Publishing, 2022, vol. 13374, pp. 95--106.

    \bibitem{appsAsynchronousMultiObjectTracking2025}
    A.~Apps, Z.~Wang, V.~Perejogin, T.~Molloy, and R.~Mahony, ``Asynchronous {{Multi-Object Tracking}} with an {{Event Camera}},'' in \emph{Proc. IEEE Int. Conf. Robot. Autom.}, Atlanta, GA, USA, 2025.

    \bibitem{wangAsynchronousBlobTracker2024}
    Z.~Wang, T.~Molloy, P.~{van Goor}, and R.~Mahony, ``Asynchronous {{Blob Tracker}} for {{Event Cameras}},'' \emph{IEEE Trans. Robot.}, vol.~40, pp. 4750--4767, 2024.

    \bibitem{ruschTrackAERRealTimeEventBased2021}
    A.~Rusch and T.~Roesgen, ``{{TrackAER}}: {{Real-Time Event-Based Particle Tracking}},'' in \emph{Proc. Int. Symp. Part. Image Velocim.}, vol.~1, no.~1, Aug. 2021.

    \bibitem{hamannETAPEventbasedTracking2025}
    F.~Hamann, D.~Gehrig, F.~Febryanto, K.~Daniilidis, and G.~Gallego, ``{{ETAP}}: {{Event-based Tracking}} of {{Any Point}},'' in \emph{Proc. IEEE/CVF Conf. Comput. Vis. Pattern Recognit.}, Nashville, TN, USA, 2025.

    \bibitem{wangEventStreambasedVisual2024a}
    X.~Wang, S.~Wang, C.~Tang, L.~Zhu, B.~Jiang, Y.~Tian, and J.~Tang, ``Event {{Stream-based Visual Object Tracking}}: {{A High-Resolution Benchmark Dataset}} and {{A Novel Baseline}},'' in \emph{Proc. IEEE/CVF Conf. Comput. Vis. Pattern Recognit.}, Seattle, WA, USA, 2024, pp. 19\,248--19\,257.

    \bibitem{zhangEfficientVisionTransformer2026}
    J.~Zhang, X.~Yang, H.~Tang, Y.~Wang, B.~Yin, H.~Wang, and X.~Fu, ``Efficient {{Vision Transformer}} with {{Token Sparsification}} for {{Event-Based Object Tracking}},'' \emph{Int. J. Comput. Vis.}, vol. 134, no.~2, p.~75, Feb. 2026.

    \bibitem{caoTOPICParallelAssociation2025}
    X.~Cao, Y.~Zheng, Y.~Yao, H.~Qin, X.~Cao, and S.~Guo, ``{{TOPIC}}: {{A Parallel Association Paradigm}} for {{Multi-Object Tracking Under Complex Motions}} and {{Diverse Scenes}},'' \emph{IEEE Trans. Image Process.}, vol.~34, pp. 743--758, 2025.

    \bibitem{liLearningGlobalObjective2022}
    S.~Li, Y.~Kong, and H.~Rezatofighi, ``Learning of {{Global Objective}} for {{Network Flow}} in {{Multi-Object Tracking}},'' in \emph{Proc. IEEE/CVF Conf. Comput. Vis. Pattern Recognit.}, 2022, pp. 8855--8865.

    \bibitem{geYOLOXExceedingYOLO2021}
    Z.~Ge, S.~Liu, F.~Wang, Z.~Li, and J.~Sun, ``{{YOLOX}}: {{Exceeding YOLO Series}} in 2021,'' \emph{arXiv preprint arXiv:2107.08430}, 2021.

    \bibitem{ballasDelvingDeeperConvolutional2016}
    N.~Ballas, L.~Yao, C.~Pal, and A.~Courville, ``Delving {{Deeper}} into {{Convolutional Networks}} for {{Learning Video Representations}},'' \emph{arXiv preprint arXiv:1511.06432}, 2016.

    \bibitem{he2020fastreid}
    L.~He, X.~Liao, W.~Liu, X.~Liu, P.~Cheng, and T.~Mei, ``Fastreid: A pytorch toolbox for general instance re-identification,'' \emph{arXiv preprint arXiv:2006.02631}, 2020.

\end{thebibliography}
\end{document}